\documentclass[lettersize,journal]{IEEEtran}
\usepackage{amsmath,amsfonts}
\usepackage{algorithmic}
\usepackage{algorithm}
\usepackage{array}
\usepackage[caption=false,font=normalsize,labelfont=sf,textfont=sf]{subfig}
\usepackage{textcomp}
\usepackage{stfloats}
\usepackage{url}
\usepackage{verbatim}
\usepackage{graphicx}
\usepackage{cite}
\usepackage[utf8]{inputenc} 
\usepackage[T1]{fontenc}    
\usepackage{hyperref}       
\usepackage{url}            
\usepackage{booktabs}       
\usepackage{amsfonts}       
\usepackage{nicefrac}       
\usepackage{microtype}      
\usepackage{xcolor}         

\usepackage{orcidlink}
\usepackage{graphicx}       
\usepackage{tabularx}  
\usepackage{pifont}
\usepackage{amsmath}
\usepackage{multirow}
\usepackage{adjustbox}
\usepackage{tikz}
\usepackage{pgfplots}
\pgfplotsset{compat=1.18}  
\usepackage{wrapfig}
\usepackage{makecell}
\usepackage[most]{tcolorbox}

\begin{document}

\title{Enhancing Visual Programming for Visual Reasoning \\ via Probabilistic Graphs}

\author{
Wentao Wan~\orcidlink{0009-0004-2063-4382},
~Kaiyu Wu~\orcidlink{0009-0002-9748-7531},
~Qingyang Ma~\orcidlink{0009-0002-0497-2395},
~Nan Kang,
~Yunjie Chen,
~Keze Wang~\orcidlink{0000-0002-7817-8306}$^{*}$,

~Liang Lin~\orcidlink{0000-0003-2248-3755},
~\IEEEmembership{Fellow,~IEEE}
\thanks{$^{*}$Keze Wang is the corresponding author.}
\thanks{
The authors are with the School of Computer Science and Engineering, Sun Yat-sen University, Guangzhou, Guangdong 510006, China
(E-mail: wanwt3@mail2.sysu.edu.cn; kezewang@gmail.com; lianglin@ieee.org).
}
}



\maketitle

\begin{abstract}
Recently, Visual Programming (VP) based on large language models (LLMs) has rapidly developed and demonstrated significant potential in complex Visual Reasoning (VR) tasks.
Previous works to enhance VP have primarily focused on improving the quality of LLM-generated visual programs.
However, they have neglected to optimize the VP-invoked pre-trained models, which serve as modules for the visual sub-tasks decomposed from the targeted tasks by VP.
The difficulty is that there are only final labels of targeted VR tasks rather than labels of sub-tasks.
Besides, the non-differentiable nature of VP impedes the direct use of efficient gradient-based optimization methods to leverage final labels for end-to-end learning of the entire VP framework.
To overcome these issues, we propose EVPG, a method to Enhance Visual Programming for visual reasoning via Probabilistic Graphs. 
Specifically, we creatively build a directed probabilistic graph according to the variable dependency relationships during the VP executing process, which reconstructs the non-differentiable VP executing process into a differentiable exact probability inference process on this directed probabilistic graph.
As a result, this enables the VP framework to utilize the final labels for efficient, gradient-based optimization in end-to-end supervised learning on targeted VR tasks.
Extensive and comprehensive experiments demonstrate the effectiveness and advantages of our EVPG, showing significant performance improvements for VP on three classical complex VR tasks: GQA, NLVRv2, and Open Images. 
\end{abstract}

\begin{IEEEkeywords}
Visual Programming, Supervised Learning, Visual Models, Visual Reasoning
\end{IEEEkeywords}

\section{Introduction}
\label{sec:intro}

Visual Reasoning (VR)~\cite{johnson2017clevr,hudson2019gqa,suhr2019corpus} is a category of complex Visual Question Answering (VQA) tasks~\cite{antol2015vqa}, which usually need a multi-step reasoning process for humans to answer. 
Existing VR solutions fall into two categories:
\textbf{pure neural} approaches~\cite{deng2024cfrnet,li2023blip,anderson2018bottom}, and \textbf{neuro-symbolic (NS)} methods~\cite{yi2018neural,sarker2021neuro,mao2019neuro}.
While pure neural methods process visual and linguistic inputs in an end-to-end fashion and achieve strong performance, they lack interpretability due to their implicit reasoning mechanisms~\cite{chakraborty2017interpretability,li2022interpretable}.
In contrast, NS approaches offer a modular pipeline: a neural module generates a symbolic reasoning program, which orchestrates visual models to perform sub-tasks such as object detection.
Then, the results are aggregated symbolically to produce the final answer.
This design aligns with human reasoning and improves transparency.

With the rise of large language models (LLMs)~\cite{zhao2023survey,yang2024harnessing}, a new NS paradigm called \textbf{Visual Programming (VP)}~\cite{gupta2023visual,suris2023vipergpt} has emerged.
VP leverages LLMs to synthesize executable visual programs for a given question.
These programs invoke pre-trained visual models to solve sub-tasks and aggregate the results for final decision-making.
VP provides strong interpretability and does not require task-specific training, making it an attractive solution for real-world VR tasks.

\begin{table}[t]
  \caption{Comparison of our EVPG with SDVP and CLOVA.}
  \label{tab:indirect}
  \centering
  \begin{tabular}{@{}cccc@{}}  
    \toprule
    \multirow{2}{*}{Method} & Don't Need & Don't Need & End-to-End \\
                            & Extra Data & Teacher Model & Optimizing \\
    \midrule
    SDVP~\cite{wan2025sdvp}     & \ding{51} & \ding{55} & \ding{55} \\
    CLOVA~\cite{gao2024clova}    & \ding{55} & \ding{51} & \ding{55} \\
    Our EVPG & \ding{51} & \ding{51} & \ding{51} \\
    \bottomrule
  \end{tabular}
\end{table}

Despite its promise, VP still lags behind pure neural models in performance, mainly due to limitations of the LLM-generated programs and the zero-shot nature of the invoked visual models~\cite{gupta2023visual}.
Recent work~\cite{khan2024self,gao2024fine} attempts to improve VP by refining program generation, or using pseudo-labels~\cite{wan2025sdvp} and reflection mechanisms~\cite{gao2024clova} to fine-tune perceptual modules.
However, these methods often rely on additional supervision, teacher models, or suffer from unreliable program correction.
Moreover, they ignore the performance impact of VP-invoked vision models, which account for the majority of the errors in our experimental observations (shown in Fig.~\ref{fig5}).

Different from previous works, we consider whether it is feasible to directly use efficient gradient-based optimization methods to leverage the original VR task labels for end-to-end fine-tuning of the entire VP framework, including the vision models it invokes.
The difficulty is that the VP framework, as shown in the top-left part of Fig.~\ref{fig_method}, is non-differentiable because it stores the argmax prediction from the distribution output of invoked visual models and may involve other non-differentiable operations.
To overcome this issue, we propose \textbf{EVPG}, a method to \textbf{E}nhance \textbf{V}isual \textbf{P}rogramming for visual reasoning via Probabilistic \textbf{G}raphs.
Tab.~\ref{tab:indirect} illustrates the advantage of our EVPG, compared to the SOTA VP-enhanced methods, i.e, SDVP~\cite{wan2025sdvp} and CLOVA~\cite{gao2024clova}.


Specifically, we have discerned that the reasoning process of any VP visual program can be represented as a directed graph according to the variable dependency relationships within that visual program. 
As shown at the top of Fig.~\ref{fig_method}, the nodes of this probabilistic graph are heterogeneous, encompassing the input/intermediate/result variables from the visual program and the weight parameters of the visual models it invokes.
The edges of the graph represent the probabilistic dependencies between these node variables.
When a variable is derived from executing a visual module within the visual program, the distribution of this node variable depends on the distribution of the input variables to that module and the weight parameters of the visual models called upon to execute the module.
In the constructed probabilistic graph, this is represented by drawing directed edges from both the input variable nodes and weight parameter nodes to the output variable node. 
Given the limited length of visual programs generated by VP when tackling VR tasks, and consequently the limited scale of the constructed probabilistic graph, we adopt the Variable Elimination~\cite{shachter1986evaluating}, an Exact Inference technique~\cite{pearl1988probabilistic} to reconstruct the originally non-differentiable execution process of VP into a differentiable process of exact probabilistic inference on the corresponding directed probabilistic graph.
As a result, this endows an efficient, gradient-based optimization for VP-invoked pre-trained visual models just utilizing the labels of the target VR tasks, thus helping the entire VP fine-tune on target VR tasks. 

To address the challenge of a vast optimizing space in the multi-step optimization process supervised by outcome labels, we further propose a curriculum learning setting.
This enables better learning performance within limited computational resources and time.
Besides, the optimizing space grows linearly with the length of the visual program during training.
Thus, we can directly arrange the curriculum learning sequence according to the step length of the visual program.

\textbf{Our contributions can be summarized as follows}:

\begin{itemize}
    \item We innovatively construct a heterogeneous directed probabilistic graph to represent the variable probability dependency relationships within the reasoning process of VP.  
    \item Utilizing the constructed probabilistic graph, we reconstruct the non-differentiable VP reasoning process into a differentiable exact probability inference process in the probabilistic graph, hence enabling the VP-invoked visual models to end-to-end fine-tune VP's sub-tasks with only original VR labels supervising. 
    \item Extensive and comprehensive experiments demonstrate the effectiveness and superiority of our EVPG, achieving 3.7\%, 7.6\%, and 20.0\% performance improvement for VP on three classical complex VR tasks: GQA~\cite{hudson2019gqa}, NVLRv2~\cite{suhr2019corpus}, and Open Images~\cite{shao2024visual}. 
\end{itemize}

\section{Related Works}
\label{related_works}

\textbf{Neuro-symbolic (NS) methods for Visual Reasoning.}
NS methods are suitable for Visual Reasoning (VR) tasks because, from a human perspective, addressing such tasks requires both visual perception and reasoning based on the perceived results.
Previous NS methods for VR usually pre-design a domain-specific language (DSL) that includes a series of predefined operations expressed by a symbolic function.
Then, NS methods utilize a sequence-to-sequence neural network to transform the original complex VR questions into a program compound of a sequence of operations in DSL.
NS-VQA~\cite{yi2018neural} utilizes a neural network to produce an attribute table from the image and then does symbolic reasoning by manipulating the table.
NS-CL~\cite{mao2019neuro} goes further and has designed a differentiable formulation for each DSL operation to endow an end-to-end optimization for the system.
D-FOL~\cite{amizadeh2020neuro} can decouple natural image perception and reasoning and uses first-order logic formulas for reasoning.
Previous NS methods rely on DSLs, which limit flexibility, and the difficulties associated with simultaneously optimizing reasoning and perception structures have restricted further development in this area. VP~\cite{gupta2023visual,suris2023vipergpt} can be regarded as a type of NS method that uses LLMs to generate code representing the visual reasoning process and uses pre-trained visual models to perform visual sub-tasks, thus achieving commendable performance on VR tasks without training.
Our work is dedicated to enhancing VP's performance on VR tasks.

\textbf{Methods for Enhancing Visual Programming.}
There are already some works that aim to enhance VP.
De-fine~\cite{gao2024fine} mainly uses a refining mechanism to better decompose the original task based on the feedback.
\cite{khan2024self} utilizes reinforcement learning to improve the LLM to generate visual programs with higher quality.
In addition to improving the quality of the visual programs, enhancing the capabilities of the invoked visual models can also boost the performance of VP on specific VR tasks.
However, the non-differentiable nature of VP makes it challenging to directly use VR labels for end-to-end fine-tuning, and we also lack ground truth for the intermediate visual sub-tasks in VP.
SDVP~\cite{wan2025sdvp} utilizes pseudo-labels provided by well-trained pure network models on visual sub-tasks in VP to sequentially distill the capabilities of the pure network models into the invoked pre-trained visual models. CLOVA~\cite{gao2024clova} uses LLMs to identify errors in the VP execution stream, then refines the visual programs or conducts further training on the additional collected data for learning visual concepts not performed well.
They both can enhance VP by improving the abilities of invoked visual models.
However, SDVP needs a well-trained network as the distillation teacher, and the learning outcomes are influenced by the quality of the pseudo-labels generated by the teacher.
For CLOVA, there is no strict guarantee that errors can be accurately identified, and reinforcing each instance of visual perception errors requires the additional collection of data, making the process relatively cumbersome.
We propose a method that reconstructs the non-differentiable reasoning process of VP into a differentiable probability-exact inference process on the probabilistic graph derived from the visual program within VP.
This allows for efficient end-to-end fine-tuning of VP on the target VR tasks using gradient-based optimization, solely utilizing the target VR task labels.
This improves the capabilities of the invoked visual models and ultimately enhances the performance of VP on the target VR tasks.

\section{Methodology}
\label{method}

\begin{figure*}[t]
\centering
\includegraphics[width=0.8\textwidth]{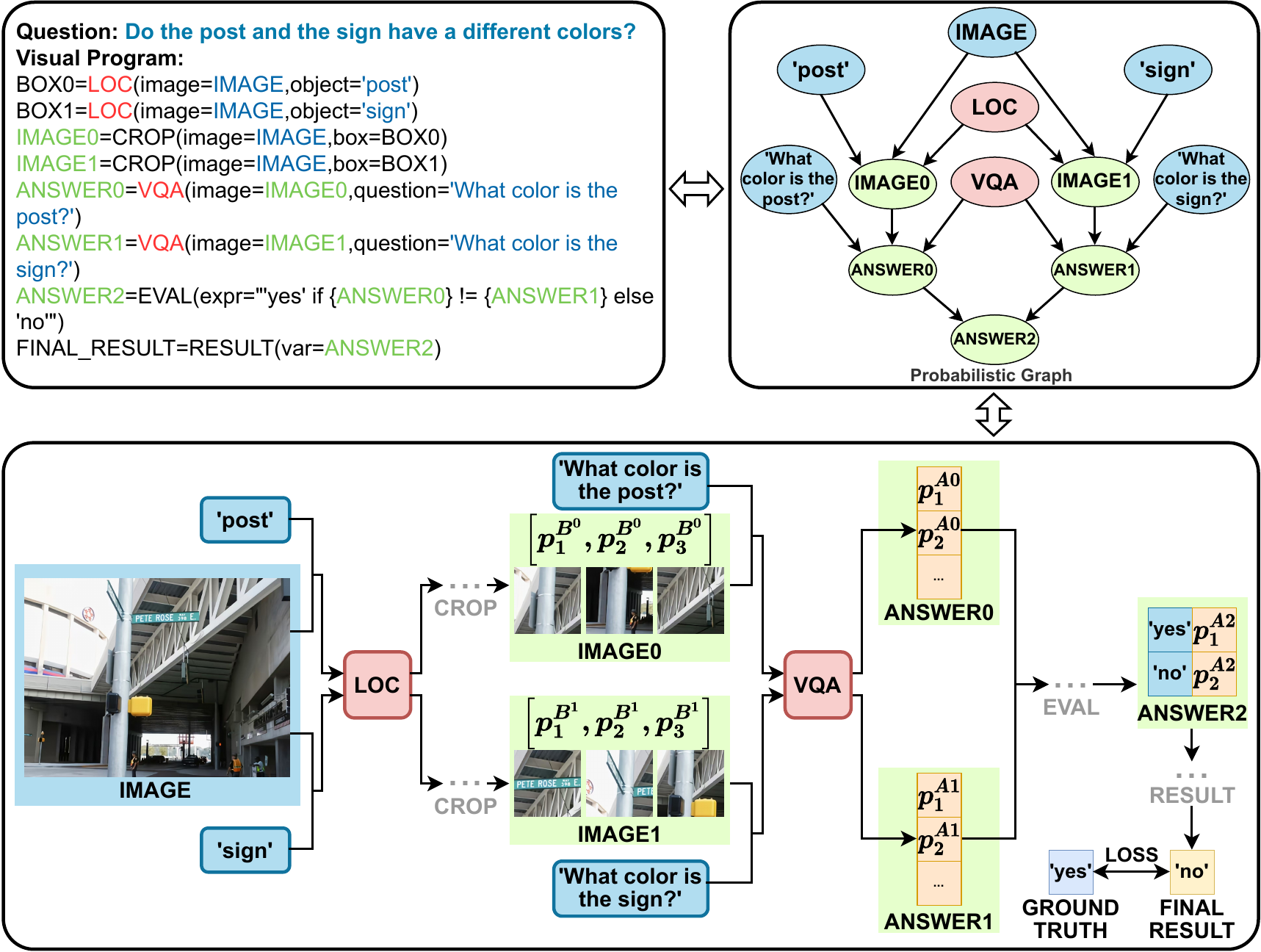}
\caption{
The overview of our proposed EVPG method.
The top-left part is the visual program, and the top-right part is the constructed Probabilistic Graph.
Below is the corresponding flowchart.
}
\label{fig_method}
\end{figure*}

In this section, we will introduce our EVPG in detail.
Firstly, we will formulate the execution process of VP.
Then, our EVPG will be introduced in three main parts: 
1) the construction of the probabilistic graph corresponding to the visual program of VP;
2) the exact inference process on the constructed probabilistic graph, which can help transform the non-differentiable visual programming execution process into a differentiable one;
3) the loss function of utilizing original VR labels supervising VP outcomes.

\subsection{Execution Process of VP}
The core idea of VP in solving VR tasks is to use LLMs to generate a visual program to address the original VR question step-by-step.
Then the visual program is executed, invoking different visual models as needed, to obtain the answer to the original VR question.

For each case in a VR task $T$, the input and label can be denoted as $<I, Q>$ and $Y$, respectively, where $I$ is the image (For GQA and Open Images, one image, NLVRv2 two images), $Q$ is the original VR question, and $Y$ is the corresponding label.
For convenience, we use VisProg~\cite{gupta2023visual}, a classical VP framework, for further introduction.
VisProg uses an LLM to accept the VR question $Q$ as input and generates a visual program $Pr$ as:
\begin{equation}
    Pr = f(Pro, Q; \theta_{llm}),
\end{equation}
where $f(\cdot;\theta_{llm})$ means an LLM with $\theta_{llm}$ as its weight parameters and $Pro$ means the prompt, which includes the instruction and a few demonstrations.
As shown in the top-left part of Fig.~\ref{fig_method}, the Visual Program $Pr$ is an executable program with multiple lines.
There are five kinds of modules in $Pr$:
LOC (the NLVR task does not have the LOC module), CROP, VQA, EVAL, and RESULT.
LOC and VQA are visual modules that invoke an object detection model and a VQA model, respectively.
The number of the LOC or VQA modules in $Pr$ could be more than one.
The process of the $i\text{-th}$ LOC module can be denoted as:
\begin{equation}
    B^i = f(I, O^i; \theta_{loc}),
\end{equation}
where $O^i$ (in Fig.~\ref{fig_method}, `post' or `sign') denotes the object name needed to be detected, $B^i$ denotes the bounding box of $O^i$ in $I$, and $f(\cdot;\theta_{loc})$ denotes the invoked object detection model which with $\theta_{loc}$ as weight parameters.
Then the sub-image $I^i$ (in Fig.~\ref{fig_method}, `IMAGE0' or `IMAGE1') is produced through:
\begin{equation}
    I^i = \text{CROP}(I, B^i).
\end{equation}
$I^i$ will then be input into the $i\text{-th}$ VQA module to produce the sub-answer $A^i$ of the sub-questions $Q^i$ in this module as:
\begin{equation}
    A^i = f(I^i, Q^i; \theta_{vqa}),
\end{equation}
where $f(\cdot; \theta_{vqa})$ is the invoked VQA model with $\theta_{vqa}$ as its weight parameters.
All $N$ ($N \geq 1$) sub-answers $\{A^i\}_{i=0}^{N-1}$ will be sent into the EVAL module to produce the final answer $A^f$ (in Fig.~\ref{fig_method}, `ANSWER2') as: 
\begin{equation}
    A^f = \text{EVAL}(\{A^i\}_{i=0}^{N-1}),
\end{equation}
where EVAL is a nested Boolean expression or a conditional selection statement that uses a nested Boolean expression as its condition.
For the formal details of EVAL, please refer to Appendix~\ref{sec: EVAL_formula}.

\subsection{Details of Our EVPG}
\label{sec: EVPG_detail}
Our goal is to enhance VP for Task $T$ by optimizing $\theta_{loc}$ and $\theta_{vqa}$ of VP-invoked pre-trained visual models in the LOC and VQA modules with original final label $Y$ as supervision.
To achieve this goal, we transform the non-differentiable execution process of the visual program $Pr$ into a differentiable probabilistic inference process on the probabilistic graph constructed from $Pr$.

\textbf{Probabilistic Graph Construction.}
We first introduce how to construct the corresponding probabilistic graph from a given visual program $Pr$.
We have observed a principle that the probability dependencies of variables in $Pr$ are directed and rely on their invoking relationship.
We argue that any given program can correspond to a specific directed probabilistic graph according to the above principle, and we refer to this relationship as ``isomorphism''.
This ``isomorphism'' provides a foundation for applying probabilistic inference tools to program variables to achieve differentiable end-to-end optimization under this explicit structure.
Here, we focus solely on end-to-end optimization for VP.

Based on that principle, we build a directed probabilistic graph, as shown in the top-right part of Fig.~\ref{fig_method}, following these steps:
First, construct a directed graph that is topologically isomorphic to the invoking structure, where each node represents the input or output variable of each module in $Pr$, and draw a directed edge from the input variable nodes to their corresponding output variable nodes;
Then, add another kind of nodes that represent the weight parameters of visual models invoked by $Pr$, and draw a directed edge from those nodes to the corresponding model output variable node, respectively. 
This represents that the probability of visual model output variables is dependent on both the model input and weights.
The constructed probabilistic graph is a heterogeneous graph, comprising three types of nodes:
input variables (blue in Fig.~\ref{fig_method}),
intermediate or final variables (green in Fig.~\ref{fig_method}),
and weight parameter variables (red in Fig.~\ref{fig_method}).
For brevity, we have omitted representations of modules such as CROP and RESULT in the probabilistic graph since they do not alter probabilities.

\textbf{Exact Inference.}
Because the visual program $Pr$ is generated from one VR question, the length of $Pr$ is limited, and thus the scale of the probabilistic graph derived from $Pr$ is limited.
Considering those, we utilize ``Variable Elimination''~\cite{shachter1986evaluating}, an Exact Inference technique~\cite{pearl1988probabilistic}, to compute and get the differentiable probability expression of the outcome nodes.
This probability serves as the answer prediction of $A^f$ for the VR question $Q$.
The expression includes the weight parameters of invoked visual models.
As the case shown in Fig.~\ref{fig_method}, from the directed probabilistic graph, when the original inputs and weight parameters (all the blue and red nodes in the generated probabilistic graph) are given as condition, the variables ${A^i}_{i \geq 0}$ are ``conditionally independent'' (in Fig.~\ref{fig_method} ANSWER0 and ANSWER1 are ``conditionally independent'').
So we can firstly calculate the probability distribution $P^{A^i}$ of each $A^i$ respectively, then utilize those distributions to calculate the distribution of the final answer prediction $A^f$ (such as ANSWER2 in Fig.~\ref{fig_method}) according to the boolean condition formulation in EVAL.
$p^{A^i}_j$ is the $j$-th component of $P^{A^i}$, represents the probability of $A^i$ on the $j$-th output candidate.
$p^{A^i}_j$ can be expressed as follows:
{\small
\begin{equation}
\begin{aligned}
\label{eq: evpg_loc_vqa1}
p^{A^i}_j &=\! P(A^i = A^i_j | I, Pr, \theta_{loc}, \theta_{vqa}) \\
              &=\! P(A^i = A^i_j | I, \{Q^i\}^{N-1}_{i=0}, \{O^i\}^{N-1}_{i=0}, \theta_{loc}, \theta_{vqa}) \\
              &=\! \! \sum_k P(A^i=A^i_j, I^i=I^i_k | I,\{Q^i\}^{N-1}_{i=0},\{O^i\}^{N-1}_{i=0}, \theta_{loc}, \theta_{vqa}) ,
\end{aligned}
\end{equation}
}
where $\{Q^i\}^{N-1}_{i=0}$ and $\{O^i\}^{N-1}_{i=0}$ respectively represent all $N$ number of sub-question $Q^i$ and object $O^i$ in $Pr$, $I^i_k$ represents the sub-image crop from the $k$-th bounding box candidate produced by the LOC model with input $(I, O^i)$.
$\theta_{loc}$ and $\theta_{vqa}$ represent the weight parameters of the LOC and VQA models, respectively.
Because variables $A^i$ and $I^i$ are not independent, we can further obtain:
{\small
\begin{equation}
\begin{aligned}
\label{eq: evpg_loc_vqa2}
p^{A^i}_j &=\! \! \sum_k P(A^i=A^i_j|I^i=I^i_k,I,\{Q^i\}^{N-1}_{i=0},\{O^i\}^{N-1}_{i=0}, \theta_{loc}, \theta_{vqa}) \\
          &  \cdot P(I^i=I^i_k|I,\{Q^i\}^{N-1}_{i=0},\{O^i\}^{N-1}_{i=0}, \theta_{loc}, \theta_{vqa}).
\end{aligned}
\end{equation}
}
By simplifying Equation~\ref{eq: evpg_loc_vqa2} according to the conditional dependencies in the probabilistic graph, we obtain that 
{\small
\begin{gather}
p^{A^i}_j = \! \! \sum_k P(A^i=A^i_j|I^i=I^i_k, Q^i, \theta_{vqa})\cdot P(I^i=I^i_k|I,O^i,\theta_{loc}),
\\
P(A^i=A^i_j|I^i=I^i_k, Q^i, \theta_{vqa}) = f_j(I^i_k, Q^i; \theta_{vqa}),
\end{gather}
}
where $f_j(I^i_k, Q^i; \theta_{vqa})$ represents the probability on the $j\text{-th}$ output candidate of the VQA model with Input $(I^i_k, Q^i)$.
Because the CROP function does not change the probability, we use $p^{B^i}_k$ representing the probability of the $k$-th bounding box candidate produced by the LOC model with weights $\theta_{loc}$ under input $(I, O^i)$ (also represented as $f_k(I, O^i; \theta_{loc})$).
So we get that:
{\small
\begin{equation}
\begin{aligned}
\label{eq: P_Ii_k}
p^{B^i}_k &= P(I^i=I^i_k | I, O^i, \theta_{loc}) \\
        &= P(B^i=B^i_k | I, O^i, \theta_{loc}) \\
        &= f_k(I, O^i; \theta_{loc}).
\end{aligned}    
\end{equation}
}
Finally, we can express $p^{A^i}_j$ as:
\begin{equation}
\label{eq:evpg_loc_vqa3}
    p^{A^i}_j = \sum_k f_j(I^i_k, Q^i; \theta_{vqa}) \cdot f_k(I, O^i; \theta_{loc}).     
\end{equation}
After we have obtained $\{P^{A^i}\}_{i=0}^{N-1}$, the differentiable probability expressions of all sub-answers, we pass them into the EVAL boolean condition formulation to get the final differentiable answer prediction distribution:
\begin{equation}
y_l = f_{eval}(\{P^{A^i}\}^{N-1}_{i=0}; \text{EVAL}(\{A^i\}^{N-1}_{i=0})),
\end{equation}
where $y_l$ represents the probability of the $l$-th candidate of VP answer prediction for VR question $Q$, $f_{eval}(\cdot; \text{EVAL})$ denotes probability calculated based on the EVAL Boolean expression. 
Please refer to Appendix~\ref{sec: EVAL_formula} for the details of $f_{eval}$. 

\textbf{Loss Function.}
After we get the probability expression $y_l$ of the final outcome, we use the cross-entropy function to construct the loss as follows: 
\begin{equation}
    L  = \frac{1}{m}\sum_m\sum_l Y_l \log y_l,
\end{equation}
where $Y_l$ denotes the label on the $l$-th dimension of the original label $Y$, which means the ground truth on this answer candidate, $m$ is the total number of training cases.
Because $y_l$ is differentiable, the loss $L$ is differentiable.
Hence, it can be trained through gradient-based optimization methods (such as SGD) to enhance the performance of VisProg on the objective VR task $T$.

\section{Experiments}

\begin{table*}[!t]
  \caption{Results of different methods on GQA, NLVRv2, and Open Images.
  Our EVPG is effective for both BLIP and InternVL1.5-2B.
  $ \text{BLIP}_{hg} $ refers to the Hugging Face~\cite{jain2022hugging} implementation of BLIP
  , and $ \text{BLIP}_{lavis} $ refers to the LAVIS~\cite{li-etal-2023-lavis} implementation of BLIP.
  There are slight performance differences between the two libraries.}
  \label{tab:main_results}
  \centering
  \begin{tabular}{@{}ccc|ccc@{}}
    \toprule
   \multicolumn{3}{c|}{\makecell{Method}} & \makecell{GQA \\ test-dev balanced} & \makecell{NLVRv2 \\ public test} & \makecell{Open Images \\ test} \\
    \midrule
     \multirow{7}{*}{VisProg}
     & \multirow{3}{*}{Zero-shot (baseline)}
     & $\text{BLIP}_{hg}$~\cite{li2022blip}       & 48.6 & 68.3 & 35.6 \\
    && $\text{BLIP}_{lavis}$~\cite{li2022blip}    & 51.0 & 73.6 & 44.9\\
    && InternVL1.5~\cite{chen2024far}            & 56.3 & 78.2 & 40.6 \\
     \cmidrule(lr){2-6} 
     & \multirow{3}{*}{Supervised}
     & $\text{BLIP}_{hg}$ (SDVP~\cite{wan2025sdvp}) & 50.4(+1.8) & 75.2(+6.9) & -\\
    && $\text{BLIP}_{lavis}$ (Our \textbf{EVPG})    & 54.7(+\textbf{3.7}) & \textbf{81.2}(+\textbf{7.6}) & 64.9(+20.0)\\
    && InternVL1.5 (Our \textbf{EVPG})            & \textbf{58.9}(+2.6) & 80.5(+2.3)& \textbf{85.1}(+\textbf{44.5})\\
    \bottomrule
  \end{tabular}
\end{table*}

\begin{table}[t]
  \caption{The settings of curriculum learning. ``Program steps'' are defined as the total number of modules that invoke the visual model.
  The other modules (e.g., CROP, EVAL, and RESULT) that do not call the visual model are excluded from the count.}
  \label{tab:train_data}
  \centering
  \begin{tabular}{@{}cccccc@{}}
    \toprule
    \small
    \multirow{2}{*}{Visual Model}  & \multicolumn{2}{c}{Curriculum Learning} & \multicolumn{2}{c}{Data Size} \\
     & stage & program steps &  GQA & NLVRv2 \\
    \midrule
    \multirow{4}{*}{BLIP} & 1 & $\leq$ 1& 15,000 & 9,681\\ 
     &  2 & $\leq$ 2 & 45,000 & 52,702\\
     &  3 & $\leq$ 3 & 60,000 & 54,866\\
     &  4 & $\leq$ 4 & 70,000 & 76,267 \\
    \bottomrule
  \end{tabular}
\end{table}

\textbf{Dataset.} We evaluate our EVPG under VisProg~\cite{gupta2023visual} (a popular VP framework) on three widely employed VR tasks: \textbf{GQA}~\cite{hudson2019gqa} and \textbf{NLVRv2}~\cite{suhr2019corpus}, along with the \textbf{Open Images} split of Visual CoT~\cite{shao2024visual} tasks (denoted as Open Images hereafter), all three require real-world visual perception and compositional reasoning abilities.
Each case in GQA and Open Images is constructed with one real-world image and a question asking about this image, while NLVRv2 is constructed with two real-world images and a statement to be determined as True or False. 
Following the prompt of VisProg (shown in Appendix~\ref{prompt_appendix}), we use DeepSeek-V3~\cite{liu2024deepseek} to generate visual programs for the cases in a subset of train balanced in GQA, and whole train splits in NLVRv2 and Open Images, which are then executed to produce the predicted answer.
We collect the programs that are successfully compiled from the training set as our training data.
Accuracy is reported respectively on the test-dev balanced of GQA, public test split of NLVRv2, and test split of Open Images.

\textbf{Implementation Details.}
Following VisProg, we use Owl-ViT-large~\cite{minderer2022simple} as the LOC model, and BLIP~\cite{li2022blip} as the VQA model.
Besides, we also evaluate our EVPG by adopting a more recent model, InternVL1.5-2B~\cite{chen2024far}, as the VQA model.
For BLIP, while learning on GQA, we unfreeze the weights of the last 4 layers of its vision encoder and answer decoder; while learning on NLVRv2, we unfreeze the last 2 layers of its vision encoder and answer decoder because we think NLVRv2, with only True or False as the answer, is easier than GQA.
For InternVL1.5-2B, we employ LoRA~\cite{hu2022lora} to fine-tune its LLM component, which helps in controlling computational resources.
We use AdamW~\cite{loshchilovdecoupled} optimizer with learning rate ${1 \times 10^{-5}}$, batch size 32 on GQA and Open Images, 64 on NLVRv2 for training.
In our experiments, we classify the difficulty of problems according to the length of program steps, and the samples with fewer steps are regarded as relatively easier ones.
We apply a four-stage Curriculum Learning (CL) to our training process, which starts with samples of 1 visual module step and gradually progresses to more challenging ones of 4 visual module steps.
The detailed curriculum learning settings are in Tab.~\ref{tab:train_data}.

\subsection{Main Results}

We compare the performance of our EVPG with an SOTA VP-enhanced method, SDVP~\cite{wan2025sdvp}, which focuses on improving VP-invoked visual models.
Note that we have not compared our EVPG with CLOVA~\cite{gao2024clova} since the size and content of the training data are very different and hard to compare fairly.
The major difference is that our EVPG optimizes a visual model invoked by an explicit multi-step process based on distant outcome supervision (original VR labels).
In contrast, CLOVA optimizes a single-step process, allowing it to optimize vision models without the need for distant supervision.
CLOVA first attempts to pinpoint the erroneous step, then gathers additional labeled internet data specifically for optimizing that step, directly supervising the model at that particular stage. 
Additionally, CLOVA only reports results based on 500 randomly sampled cases, whereas we have tested our EVPG under a more general setup, evaluating its performance across the entire GQA test-dev balanced set, NLVRv2, and Open Images public test set.
From the results reported by CLOVA on 500 random samples, the performance on NLVRv2 is significantly lower than Baseline in Tab.~\ref{tab:main_results} (65.6 vs 73.6).

\begin{figure}
\centering
\footnotesize
\begin{tikzpicture}
\begin{axis}[
    width=4.8cm, 
    height=4cm, 
    xlabel={Curriculum Learning Stage}, 
    ylabel={Accuracy (\%)}, 
    grid=major, 
    xmin=0, xmax=4, 
    ymin=51, ymax=55, 
    xtick={0,1,2,3,4}, 
    ytick={51,52,53,54,55}, 
    legend pos=north west, 
    smooth, 
    mark=* 
]

\addplot[
    color=blue,
    mark=square*,
    line width=1pt
] coordinates {
     (0, 51.0)
    (1, 53.2)
    (2, 54.1)
    (3, 54.7)
    (4, 54.4)
};

\end{axis}
\end{tikzpicture}
\caption{The accuracy of each curriculum learning stage while learning on GQA with our EVPG.}
\label{fig: step_acc}
\end{figure}
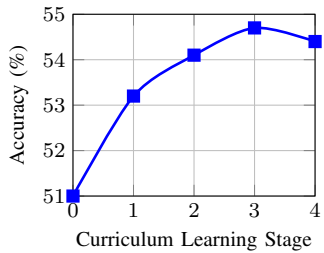

Comparison results on the three tasks are shown in Tab.~\ref{tab:main_results}. 
For the supervised methods under VisProg, we get answers by executing visual programs.
EVPG with BLIP achieves improvements on GQA (+3.7\%), NLVRv2 (+7.6\%), and Open Images (+20.0\%).
SDVP achieves an increase of 1.8\% and 6.9\% on GQA and NLVRv2, respectively. 
Compared to SDVP, our method shows a greater improvement over the baseline, which validates the advantages of our method: achieving a larger performance gain without the need for an additional teacher model.
Moreover, our EVPG remains effective when applied to the more powerful visual models.
InternVL1.5-2B obtains a 2.6\% improvement on GQA, a 2.3\% improvement on NLVRv2, and a 44.5\% improvement on Open Images, demonstrating the adaptability of our EVPG to the model.

\subsection{Ablation Studies and Robustness Analysis}

\begin{table*}[t]
  \caption{
      Cross-task evaluation: the model trained on GQA demonstrates improved accuracy when evaluated on Open Images, indicating enhanced cross-task generalization.
  }
  \centering
  \begin{tabular}{c c c | c c}
    \toprule
        \multicolumn{3}{c|}{\makecell{Method}} & \makecell{GQA\\(in-task)} & \makecell{Open Images\\(cross-task)} \\
    \midrule
        \multirow{2}{*}{End-to-end} 
            & Zero-shot  & InternVL1.5~\cite{chen2024far} & 63.2  & 41.0 \\
            & Supervised & InternVL1.5 (SFT)          & 64.6(+1.1)    & 42.7(+1.7) \\
    \midrule
        \multirow{3}{*}{VisProg} 
            & Zero-shot  & InternVL1.5~\cite{chen2024far} & 56.3  & 40.6 \\
            & \multirow{2}{*}{Supervised}  & InternVL1.5 (SDVP~\cite{wan2025sdvp}) & 57.7(+1.4)  & 42.5(+1.9) \\
            &           & InternVL1.5 (Our \textbf{EVPG}) & 58.0(\textbf{+1.7})  & \textbf{44.7(+4.1)} \\
    \bottomrule
  \end{tabular}
  \label{tab:ablation_cross_task}
\end{table*}

\begin{table*}[t]
    \begin{minipage}[t]{0.48\textwidth}
        \centering
        \caption{Results comparison with different training paradigms. ``CL'' means ``Curriculum Learning''.}
        \label{tab:ablation_training_paradigm}
        \vspace{0.5pt}
        \renewcommand{\arraystretch}{1.1}
        \begin{tabular}{@{}cccc@{}}
            \toprule
            \multirow{2}{*}{Model}  & \multirow{2}{*}{CL} & GQA & NLVRv2\\
              & & test-dev balanced & public test \\
            \midrule
            \multirow{3}{*}{BLIP} & Baseline & 51.0 & 73.6 \\
              & \checkmark & 54.7(+3.7) & 81.2(+7.6)\\
              & \ding{55} & 53.3(+2.3) & 79.3(+5.7)\\
            \bottomrule
        \end{tabular}
    \end{minipage}
    \hfill 
    \begin{minipage}[t]{0.48\textwidth}
        \centering
        \caption{Effect of Disrupted Visual Programs Proportion on BLIP Model Performance. The batch size for NLVRv2 and GQA is \textbf{32}.}
        \label{tab:ablation_noisy_data}
        \renewcommand{\arraystretch}{0.9}
        \setlength{\tabcolsep}{10pt}
        \begin{tabular}{ccc@{}}
            \toprule
            Disrupted   & GQA & NLVRv2\\
            Programs    & test-dev balanced & public test \\
            \midrule
            Baseline    &    51.0       &    73.6       \\
            0\%         &    54.7(+3.7)   &    81.8(+8.2) \\
            20\%        &    54.6(+3.6)   &    80.9(+7.3)   \\
            50\%        &    53.6(+2.6)   &    77.8(+4.2)   \\
            \bottomrule
        \end{tabular}
    \end{minipage}
\end{table*}

\textbf{Training Paradigms.}
Initially, our experiments show that training the model with all data simultaneously barely enhances its performance, likely due to the need to handle complex task variations from the start, which can destabilize training before simple patterns are learned. 

Thus, we adopt curriculum learning, starting with simpler tasks and progressively introducing more complex ones. 
We organize the training data by the number of visual program steps, beginning with one-step visual programs (only calculating steps invoking visual models), then moving to two-step programs, and gradually increasing the complexity of the training tasks.
The experiment results in Tab.~\ref{tab:ablation_training_paradigm} indicate that using a curriculum learning approach can unleash the potential of our method.
Without curriculum learning, less improvement in model performance is observed.
With curriculum learning, our EVPG shows a 3.7\% increase on GQA and a 7.6\% increase on NLVRv2.
In Fig.~\ref{fig: step_acc}, we present the performance enhancements of VP across different stages of CL when trained on GQA under our EVPG.
The effectiveness of VP reaches saturation by the third stage, suggesting that a total of 3 to 4 steps is an appropriate arrangement for the curriculum learning stage of our EVPG.

\textbf{Robustness to Noisy Visual Program.}
Our EVPG is an optimization method based on an explicit workflow, which is determined by visual programs generated by LLMs.
Therefore, the quality of the visual program has a crucial impact on the performance of our method.
However, the programs generated by LLMs can not be perfectly accurate, which necessitates that our method be robust to noisy data.
To evaluate the robustness of our method against the noise, we introduce noise into the training dataset by intentionally disrupting the visual programs generated by LLMs.
The types of disruptions include: 
1) randomly replacing module input parameters and 
2) replacing certain modules with alternatives.
It is worth mentioning that these disruptions will not cause program execution workflow errors.
We randomly select x\% of the data for disruptions.
As shown in Tab.~\ref{tab:ablation_noisy_data}, our method performs relatively well even when 20\% of the training data is corrupted, with only a 0.1\% drop in performance in GQA compared to that where no visual program is corrupted.
Only when the artificial corruption rate rises to 50\%, the performance of our EVPG drops from our best performance, with a 1.1\% decrease in GQA and a 4.0\% decrease in NLVRv2.
These results demonstrate that our EVPG has robustness against the noise of the visual program.
Moreover, combining Tab.~\ref{tab:ablation_noisy_data} and the error analysis in \cite{gupta2023visual} and Sec.~\ref{qualitative_analysis}, it is evident that current LLMs meet the quality requirement of generating visual programs.

\begin{figure*}[t]
    \centering
	\includegraphics[width=0.73\textwidth]{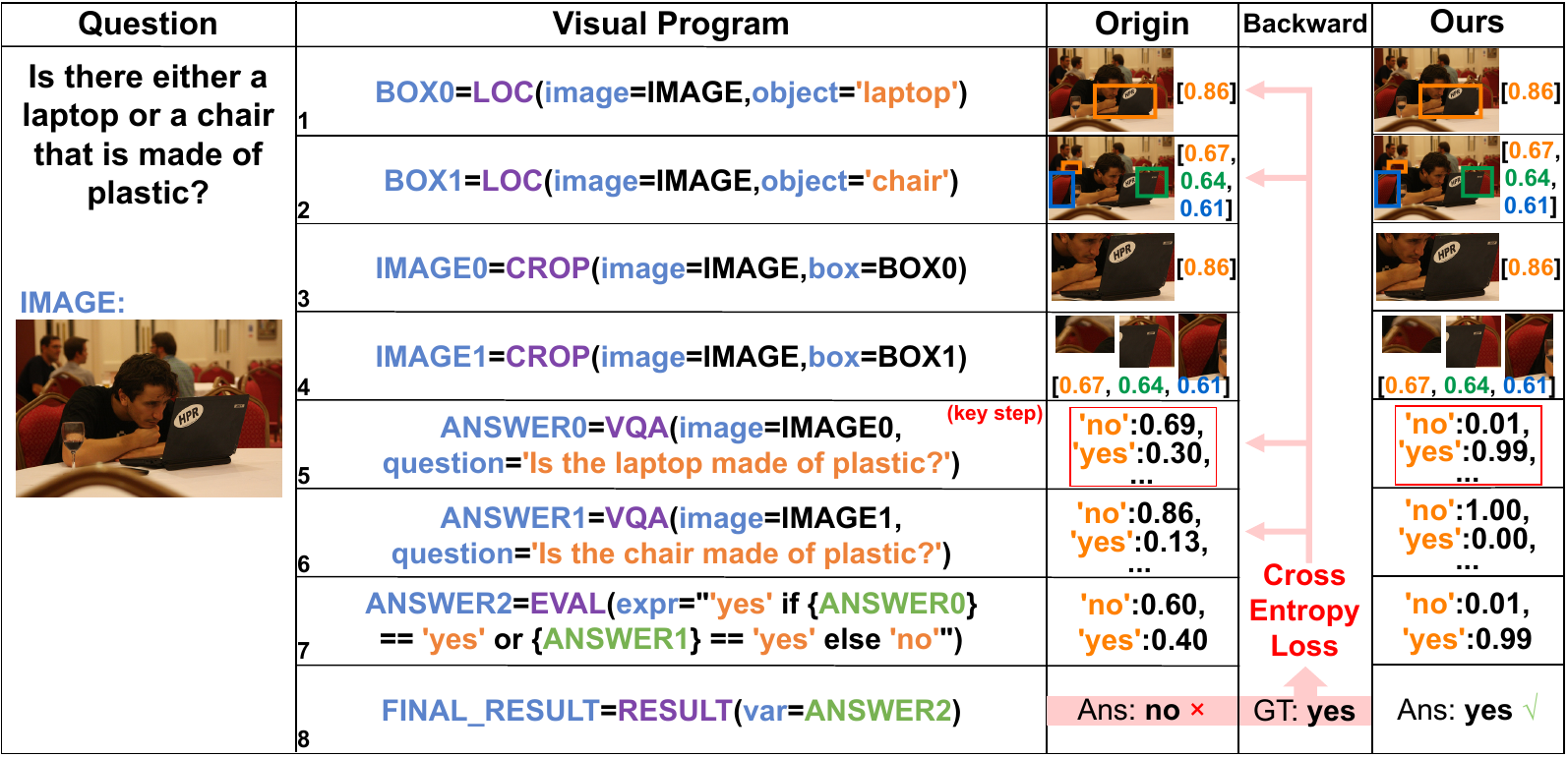}
    \caption{Qualitative example showing the program execution flow of VisProg original and after learning with our method. We provide detailed process outputs of each step during program execution. Ans means the final answer, and GT means the ground truth. For more cases, please refer to our supplementary materials.}
    \label{fig_case}
\end{figure*}

\begin{figure}[t]
\centering
    \includegraphics[width=1\columnwidth]{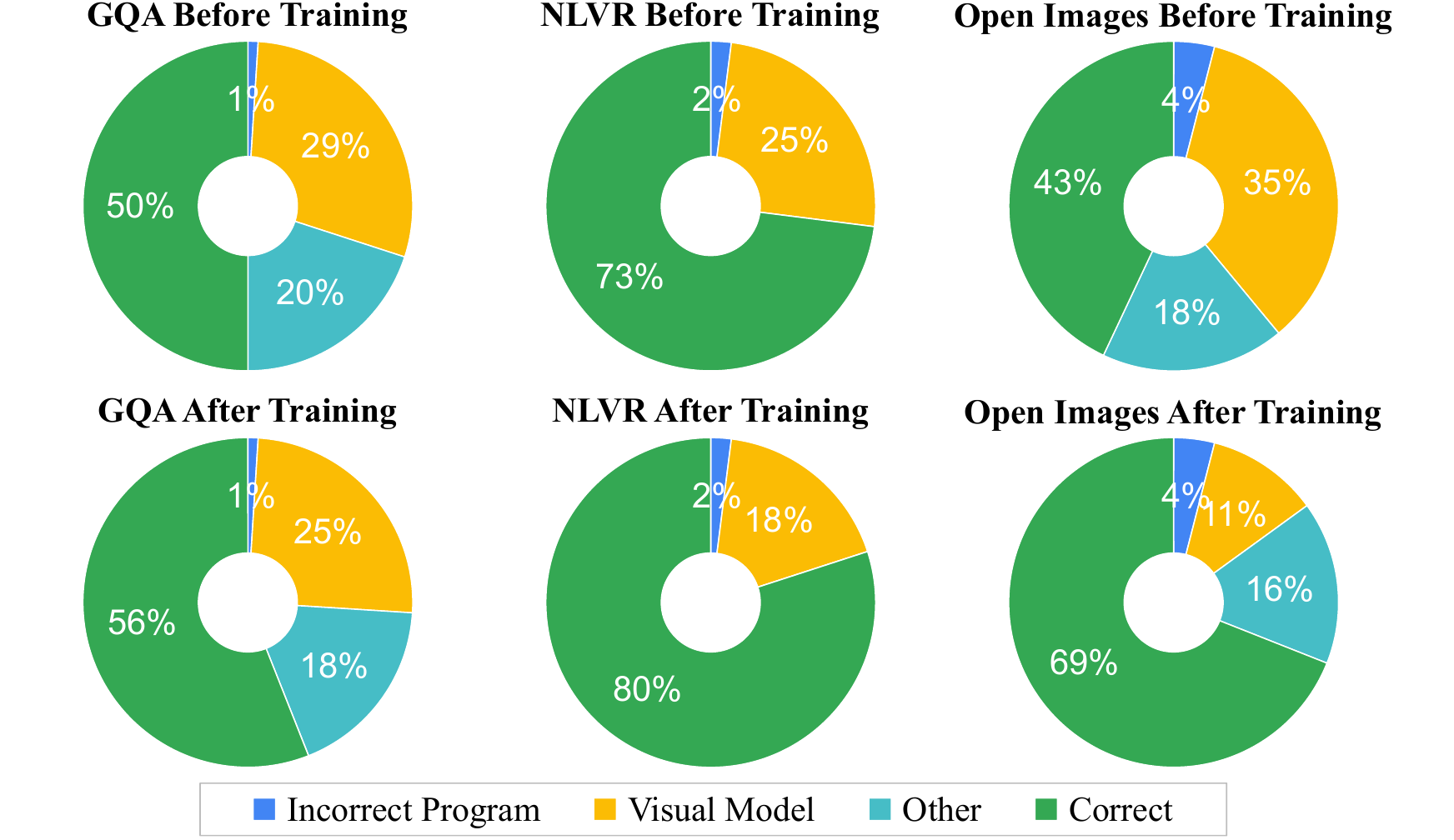}
\caption{Sources of errors in VisProg before and after EVPG. We categorize the error causes into three groups: (1) \textcolor{blue}{Incorrect Program}, (2) incorrect outputs from \textcolor{orange}{Visual Models}, and (3) \textcolor{cyan}{Other} factors such as semantically similar answers that fail to match the ground truth. 
}
\vspace{-10pt}
\label{fig5}
\end{figure}

\subsection{Analysis of Generalization and Adaptability}

\textbf{Cross-task generalization ability.}
To evaluate the generalization ability of our EVPG, we conduct a cross-task evaluation by training the model on the GQA train split and evaluating it on the Open Images test set.
As shown in Tab.~\ref{tab:ablation_cross_task}, our EVPG achieves a  4.1\% improvement in cross-task test accuracy, outperforming the 1.7\% gain obtained with the End-to-end (without VisProg) Supervised Fine-Tuning (SFT) method.
It can be observed that the performance improvement of SDVP~\cite{wan2025sdvp} (another VP-enhanced method) on GQA after training (+1.4\%) is lower than the improvement achieved by our EVPG.
Additionally, SDVP's cross-task generalization ability on Open Images (+1.9\%) is also lower than our EVPG's improvement.
We attribute this difference to the fact that our EVPG makes better use of the dataset's result labels for learning, effectively avoiding the noise introduced by SDVP during the intermediate step supervision from the teacher model.
These results demonstrate the solid cross-task generalization ability of our EVPG. 

\textbf{Long program capability.}
To demonstrate that EVPG can handle VP optimization for longer programs, we specifically select long programs from the NLVRv2 dataset as the training set.
These programs involve 6–8 steps for calling the visual model and a total program length of 8–13 steps, which is 2–3 times the average program length in the dataset.
This subset contains approximately 8,000 samples, accounting for about 10\% of the NLVRv2 training set.
Then, we train BLIP on this subset, aligning the setting in Tab.~\ref{tab:main_results}.
Our experiments show that using this long program subset can improve the VP framework's accuracy on NLVRv2 from 73.6\% to 78.3\% (+4.7\%).
We believe the performance gain is smaller than the +7.6\% in Tab.~\ref{tab:main_results} because the size of the training set in this case is only 1/10 of the full dataset.
These results demonstrate that EVPG can still enhance the performance of the VP framework even for longer programs.

\textbf{Adaptability to newer models.}
To demonstrate the adaptability of EVPG to newer models, we integrate and experiment with newer visual models on GQA.
Specifically, we use LLMDet Swin-L~\cite{fu2025llmdet} as the LOC model and InternVL3-2B~\cite{zhu2025internvl3} as the VQA model.
These models are first retrained on GQA, consistent with the settings in Tab.~\ref{tab:ablation_cross_task}.
Upon applying our EVPG, the accuracy of the VP framework utilizing these new models on the GQA test-dev balanced split increases from 56.2\% to 58.8\%.
This absolute improvement of 2.6\%, comparable to the gains observed with the models in Tab.~\ref{tab:ablation_cross_task} (Owl-ViT-large~\cite{minderer2022simple} and InternVL1.5-2B~\cite{chen2024far}), firmly validates the effectiveness of our EVPG when incorporating newer models within the VP framework.
It is worth noting that even with these updated, cutting-edge models, the baseline performance of the VP framework on GQA does not inherently increase.
However, our EVPG consistently provides a substantial performance lift, underscoring its robust adaptability and practical utility across newer vision models.

\subsection{Qualitative Analysis}
\label{qualitative_analysis}

\textbf{Error Analysis.}
To evaluate the impact of our method on visual model performance in VisProg, we randomly select 100 examples from the GQA, NLVRv2, and Open Images datasets for error analysis, respectively.
Manual inspection reveals that the primary errors in VisProg originate from visual models, and there are very few incorrect program errors, which shows that the quality of LLM-generated visual programs is already high.
These results illustrate the importance of optimizing the VP-invoked vision models.
As shown in Fig.~\ref{fig5}, for GQA, optimizing with EPVG reduces the error rate by 6\%, enhancing overall accuracy by the same margin.
Similar results are observed in the NLVRv2 and Open Images experiments, where visual model errors decrease by 7\% and 24\%, improving accuracy by 7\% and 26\%.

\textbf{Case Study.}
In Fig.~\ref{fig_case}, we visualize a case on GQA to illustrate the effectiveness of our EVPG.
The case in Fig.~\ref{fig_case} is the successful case of our EVPG on GQA.
In this case, facing the image and the question ``Is there either a laptop or a chair that is made of plastic?'', we first use the visual model LOC to find the laptop and chair, and crop them for further visual sub-questions.
Then we call the visual model VQA to learn whether the laptop is made of plastic, as well as the chair.
However, when answering the intermediate question `` Is the laptop made of plastic'', the VQA model gives a wrong rank for answers, 0.69 for ``no'' and 0.30 for ``yes'', which causes an incorrect answer  ``no'' for the intermediate question and leads to the wrong final answer.
After being optimized with our EVPG, the VQA model provides the correct answer ranking in the key step with the probability of ``yes'' at 0.99, ranked in the top 1, and we get the answer to the original question ``yes'', which is correct.
For more cases, please refer to the Appendix~\ref{sec: more_cases}.

\subsection{Why Learning under VisProg?}

Our EVPG is fundamentally a technique for optimizing neural network models invoked under an explicit procedure (here, the VisProg's visual program).
Its key contribution lies in enabling end-to-end learning with only outcome supervision for non-differentiable yet interpretable VP frameworks.
Compared to conventional end-to-end fine-tuning of pure neural networks, our EVPG is combined with explicit symbolic computation or reasoning, which preserves the interpretability advantage of visual programming.

Moreover, for compositional tasks where interfaces between the task and invoked visual modules mismatch (e.g., NLVRv2 takes two images as input, while its invoked VQA module processes only one), EVPG still permits optimization.
This expands the scope of optimizable data to broader compositional task scenarios.

We further argue that the strategy of our EVPG, which learns under explicit visual programming, offers another advantage over pure neural network paradigms: stronger cross-task generalization.
When decomposing complex vision tasks that involve intricate images and comprehension demands via visual programs, the resulting visual subtasks for model training become simpler, which contain localized images and partial understanding requirements.
This narrows down the input distribution gaps of subtasks across different tasks, allowing visual models trained on one task to generalize better to others through explicit programmatic composition.

To evaluate the advantage of our EVPG for cross-task generalization, we train visual models on the GQA~\cite{hudson2019gqa} train set using both our EVPG within the VisProg~\cite{wan2023visualprog} framework and a traditional end-to-end supervised fine-tuning approach based on pure neural networks (marked with ``SFT'').
We then evaluate their cross-task performance on the Open Images~\cite{shao2024visual} test split.
As shown in Tab.~\ref{tab:ablation_cross_task}, our EVPG achieves an accuracy of 44.7\% on Open Images, representing a 4.1\% improvement over the baseline and outperforming the pure-network-training method, which achieves 42.7\% on Open Images, only a 1.7\% improvement.
Additionally, we examine a hybrid approach ``H\_SFT'' with the VisProg, invoking the visual model fine-tuned in a pure-network end-to-end supervised learning way.
The ``H\_SFT'' method achieves an accuracy of 43.5\% under the same testing conditions on Open Images, an improvement of 2.9\%, better than the ``SFT'' setting, yet still fell behind the performance of our EVPG.
These results demonstrate that: i) evaluating with VisProg, which decomposes the original task, can improve the cross-task generalization performance; ii) training under VisProg can further improve that performance, which supports the cross-task generalization advantage of our method.

\begin{figure}[t]
  \centering
  \includegraphics[width=1\linewidth]{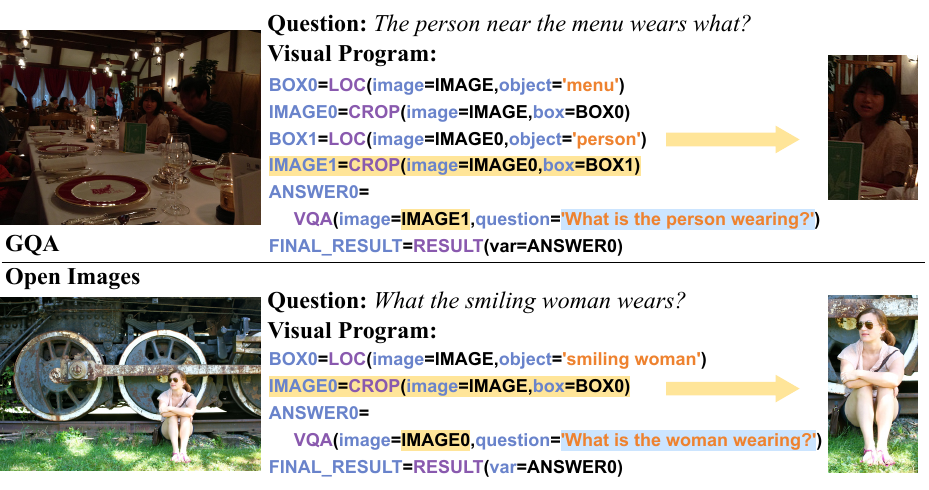}
  \caption{Example of sub-task module generalization. The top example is from the in-task GQA dataset, and the bottom is from the cross-task Open Images dataset. The VQA module learns to perform the sub-task “What is the person wearing” during GQA training, which generalizes naturally to similar queries such as “What is the woman wearing” in Open Images. Although the original images and questions differ significantly, the cropped regions after LOC and CROP operations (shown on the right) are more visually similar, which we think will help reduce input distribution shift and enhance cross-task generalization.}
  \label{fig:generalization_vqa}
\end{figure}

\begin{figure}[t]
  \centering
  \includegraphics[width=1\linewidth]{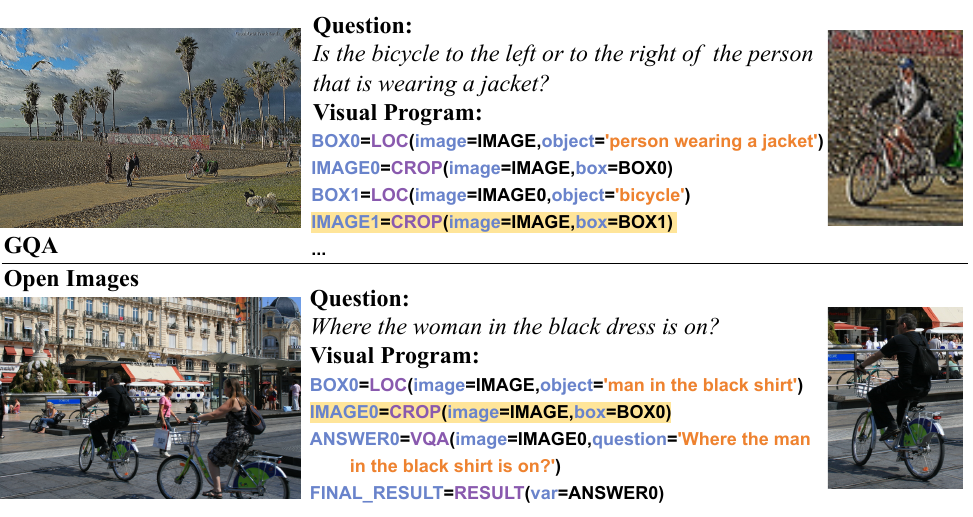}
  \caption{Example of visual input gap comparing. The top example is from the in-task GQA dataset, and the bottom is from the cross-task Open Images dataset. Although the original images and questions differ significantly, the cropped regions after LOC and CROP operations (shown on the right) are more visually similar, which we think will help reduce input distribution shift and enhance cross-task generalization.}
  \label{fig:generalization_image}
\end{figure}

We further provide a visual explanation about why learning under VisProg will benefit cross-task generalization through specific cases.
For example, in one GQA training case illustrated in Fig.~\ref{fig:generalization_vqa}, the VQA sub-module within the visual program learns to answer questions such as “What is the person wearing?”
This capability is easier to transfer to similar questions like “What is the woman wearing” in the Open Images than if learning with the original question as input (from ``The person near the menu wears what'' to ``What the smiling woman wears'').
Furthermore, the LOC and CROP operations applied to images help narrow down the input distributions between GQA and Open Images by focusing on consistent image regions of interest.
As shown on the right side of Fig.~\ref{fig:generalization_vqa} and Fig.~\ref{fig:generalization_image}, although the original images and questions in GQA and Open Images differ substantially, the localized and cropped subimages exhibit greater visual similarity, which aids in improving the cross-task generalization.

In contrast, the traditional end-to-end pure-network learning paradigm lacks explicit task structure modeling, which forces neural network models to learn mapping relationships directly from raw input-output distributions.
Due to the compositional nature of the real-world data, the distribution of the original task usually exhibits greater divergence compared to that of sub-tasks in VisProg.
Consequently, pure end-to-end learning with neural networks demonstrates relatively weaker cross-task generalization capability.

\section{Conclusion}
\label{conclusion}

In this study, we have introduced the EVPG method to improve the VP framework for specific VR tasks.
We develop a directed probabilistic graph from the visual program and implement exact inference to make the reasoning process differentiable.
This allows only utilizing the original labels of VR tasks for supervision, enabling an end-to-end optimization of the weight parameters of VP-invoked visual models to enhance the performance of VP on these tasks.
Future research will extend our EVPG to additional tasks beyond visual reasoning.

{
\appendix

\subsection{Formulation of ``EVAL''}
\label{sec: EVAL_formula}

In VisProg~\cite{gupta2023visual}, one kind of Visual Programming (VP) framework, after generating the answers to each sub-question, an EVAL module is used to combine those answers.
We can express it as:
{\small
\begin{equation}
    A^f = \text{EVAL}(\{A^i\}^{N-1}_{i=0}),
\end{equation}
}
where $A^f$ is the final answer produced by VisProg, $\{A^i\}^{N-1}_{i=0}$ represents the set of answers to sub-questions.
EVAL is a nested Boolean expression or a conditional selection statement that uses a nested Boolean expression as its condition.
The nested Boolean expression can be expressed as:
{\small
\begin{equation}
    D^f = Bool(\{D^i\}^{N-1}_{i=0}),
\end{equation}
}
where $D^f$ is the final Boolean value from which to determine $A^f$, $\{D^i\}^{N-1}_{i=0}$ is a set containing multiple atomic Boolean expressions $D^i$, each $D^i$ use $A^i$ as a variable to calculate its Boolean value.
When using our EVPG, the answer to each sub-question is represented as a probability distribution $P^{A_i}$ on a candidate answer list.
Then, we can get
{\small
\begin{equation}
\begin{aligned}
    y_l &= f_{eval}(\{P^{A^i}\}^{N-1}_{i=0}; \text{EVAL}(\{A^i\}^{N-1}_{i=0})) \\
        &= f_{eval}(\{P^{A^i}\}^{N-1}_{i=0}; Bool(\{D^i\}^{N-1}_{i=0})).
\end{aligned}
\end{equation}
}
Here, we introduce the formulation of the EVAL function to combine those distributions into the probability distribution of the final answer.
First, it is easy to calculate the Boolean probability distribution of each atomic Boolean expression $D^i$ from $P^{A^i}$ to get $P(D^i=T)$ and $P(D^i=F)$, in which `T' represents `True' and `F' represents `False'.
There are 4 ways to combine the atomic Boolean expression $D^i$ and $D^s$ into the higher-level Boolean expression $D$:
{\small
\begin{equation}
    D = \begin{cases}
        ^{\neg} D^i \\
        D^i\; \text{and} \; D^s \\
        D^i\; \text{or} \; D^s \\
        D^i\; \text{xor} \; D^s
    \end{cases}.
\end{equation}
}
For $D=^{\neg} D^i$,
{\small
\begin{equation}
    \begin{aligned}
    P(D=T) = P(^{\neg} D^i=T) = P(D^i=F) \\
    P(D=F) = P(^{\neg} D^i=F) = P(D^i=T).
    \end{aligned}
\end{equation}
}
For $D=D^i\; \text{and} \; D^s$,
{\small
\begin{equation}
    \begin{aligned}
        P(D=T) &= P(D^i=T) \cdot P(D^s=T) \\
        P(D=F) &= 1 - P(D=T).
    \end{aligned}
\end{equation}
}
For $D=D^i\; \text{or} \; D^s$,
{\small
\begin{equation}
    \begin{aligned}
        P(D=T) &= 1 - P(D=F) \\
        P(D=F) &= P(D^i=F) \cdot P(D^s=F).
    \end{aligned}
\end{equation}
}
For $D=D^i\; \text{xor} \; D^s$,
{\small
\begin{equation}
    \begin{aligned}
        P(D=T) &= P(D^i=T) \cdot P(D^s=F) + P(D^i=F) \cdot P(D^s=T) \\
        P(D=F) &= 1 - P(D=T).
    \end{aligned}
\end{equation}
}
By repeatedly applying the above formula and calculating sequentially, we can ultimately obtain the probability distribution of the Boolean value for $D^f$, from which the probability distribution $y$ of the final answer $A^f$ for the original Visual Reasoning(VR) input $<I, Q>$ can be derived.
Then we can use this answer prediction probability distribution $y$ to construct the loss function with the label of this case (please refer to Sec.~\ref{sec: EVPG_detail}).

\subsection{More Cases}
\label{sec: more_cases}

\begin{figure}[t]
	\centering
	\includegraphics[width=1\linewidth]{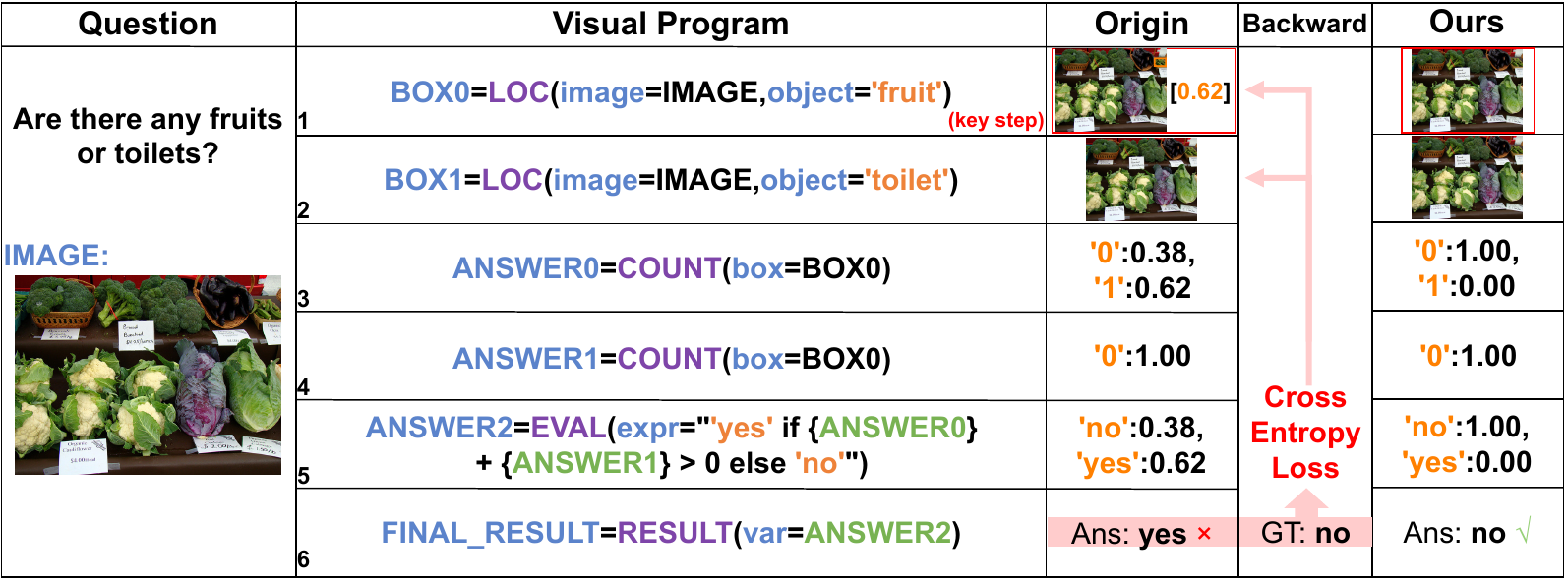}
	\\[3mm] 
	\includegraphics[width=1\linewidth]{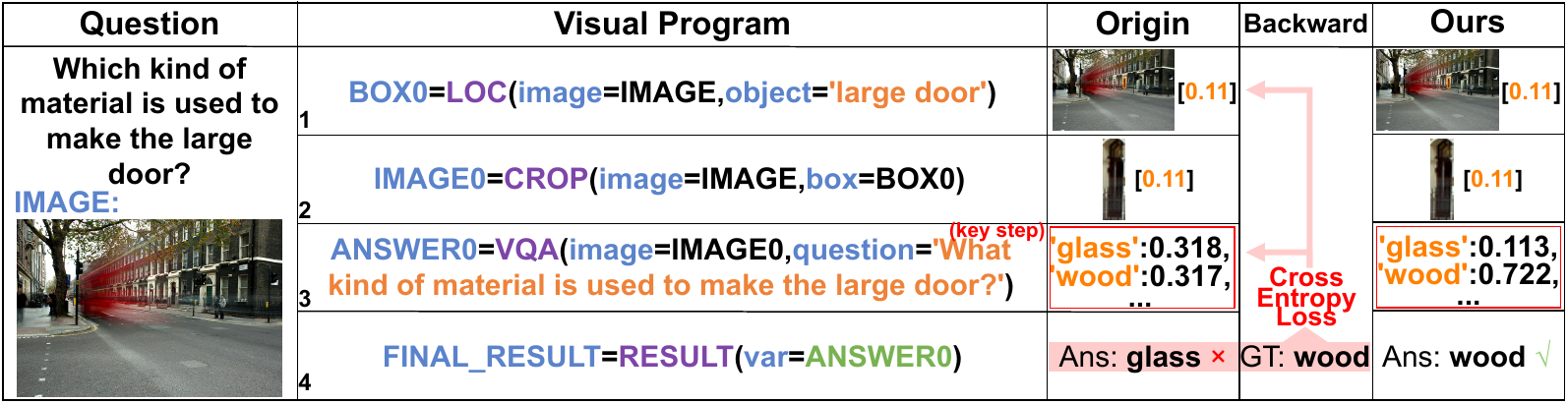}
    \caption{These program execution processes of the Visual Program before and after learning with our EVPG for a case in GQA. We provide detailed process outputs of each step during program execution. `Ans' means the final answer, and `GT' means the ground truth.}
    \label{fig_case_combined_gqa} 
\end{figure}

\begin{figure}[t]
	\centering
	\includegraphics[width=1\linewidth]{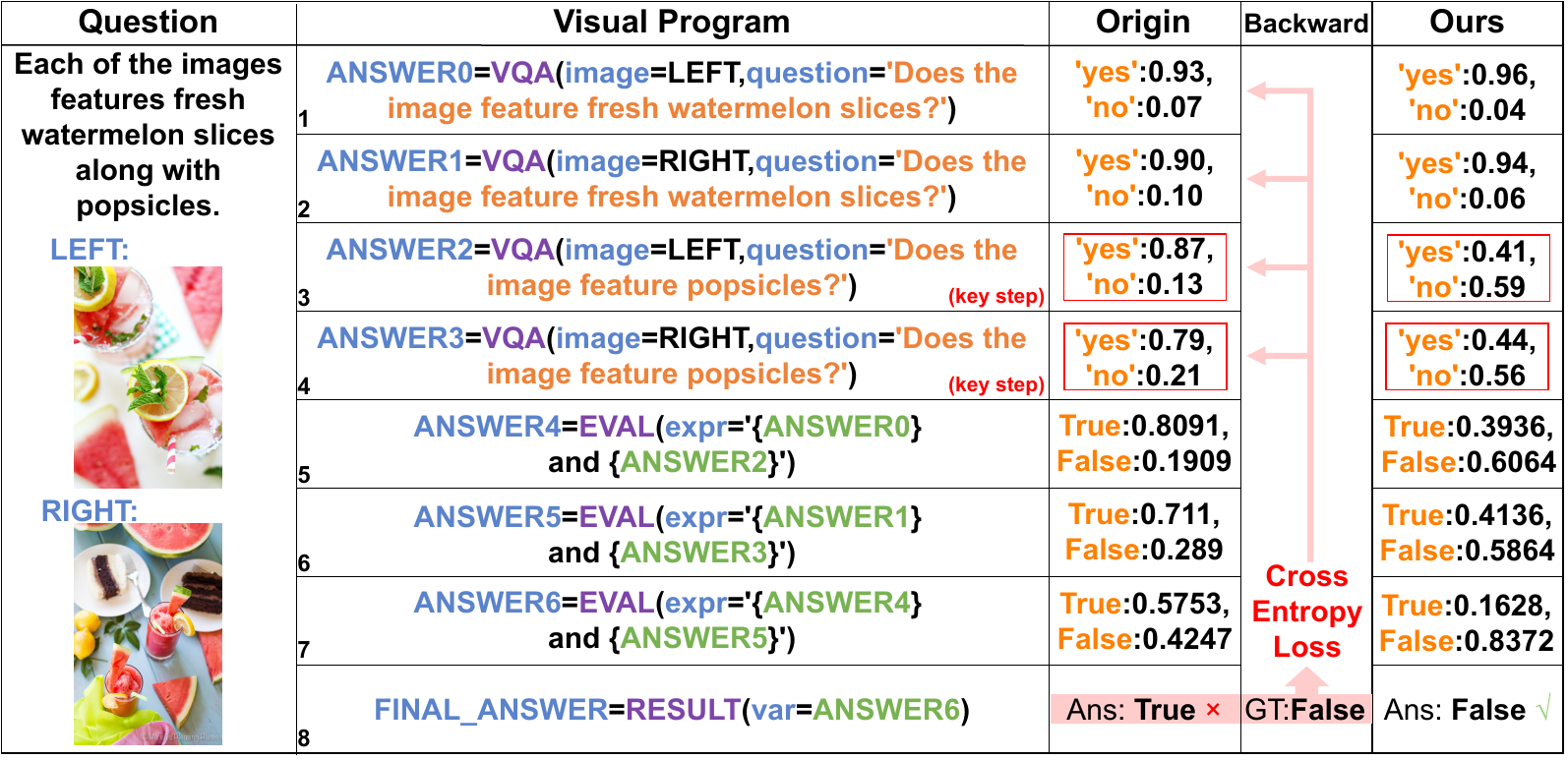}
    \\[3mm] 
    \includegraphics[width=1\linewidth]{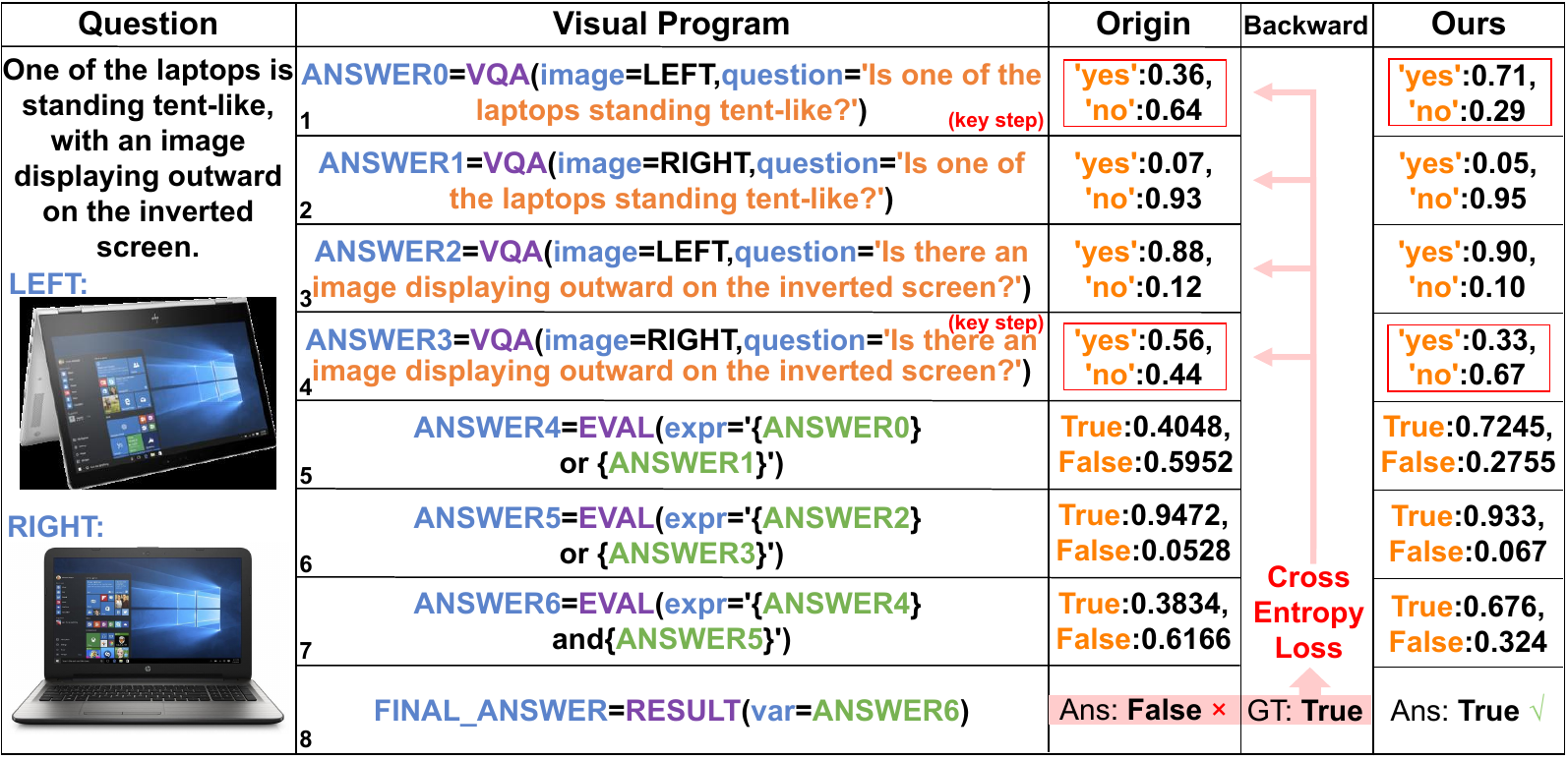}
    \caption{These program execution processes of the Visual Program before and after learning with our EVPG for a case in NLVRv2. We provide detailed process outputs of each step during program execution. `Ans' means the final answer, and `GT' means the ground truth.}
    \label{fig_case_combined_nlvr}
\end{figure}

\begin{figure}[t]
	\centering
	\includegraphics[width=1\linewidth]{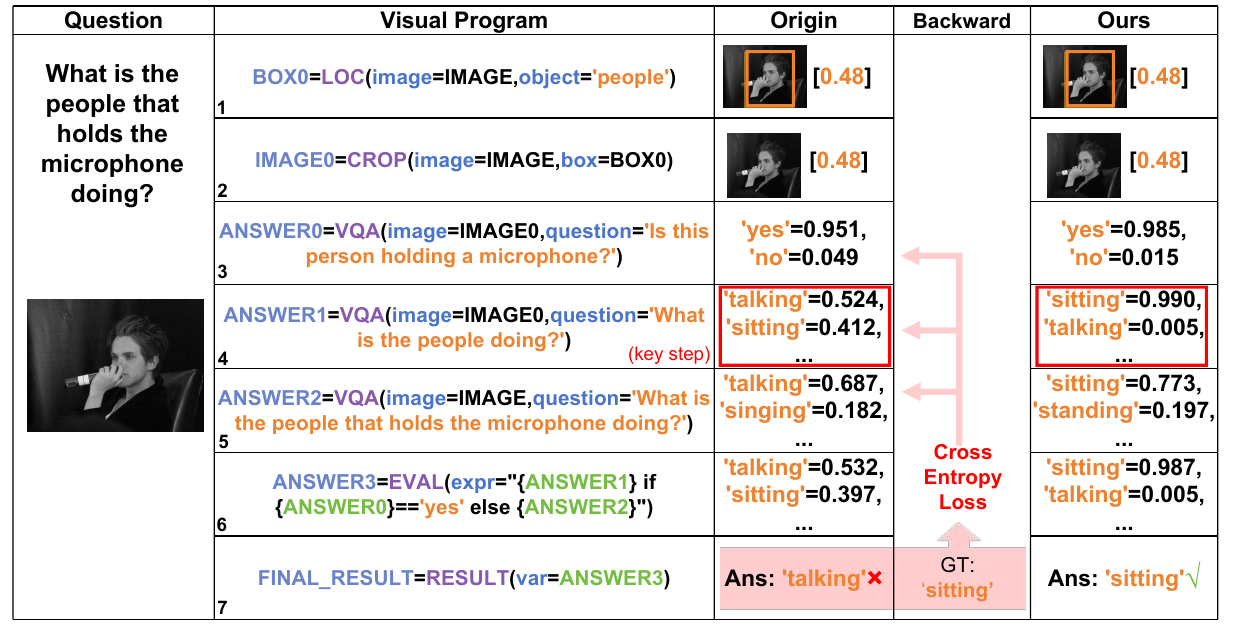}
    \\[3mm] 
    \includegraphics[width=1\linewidth]{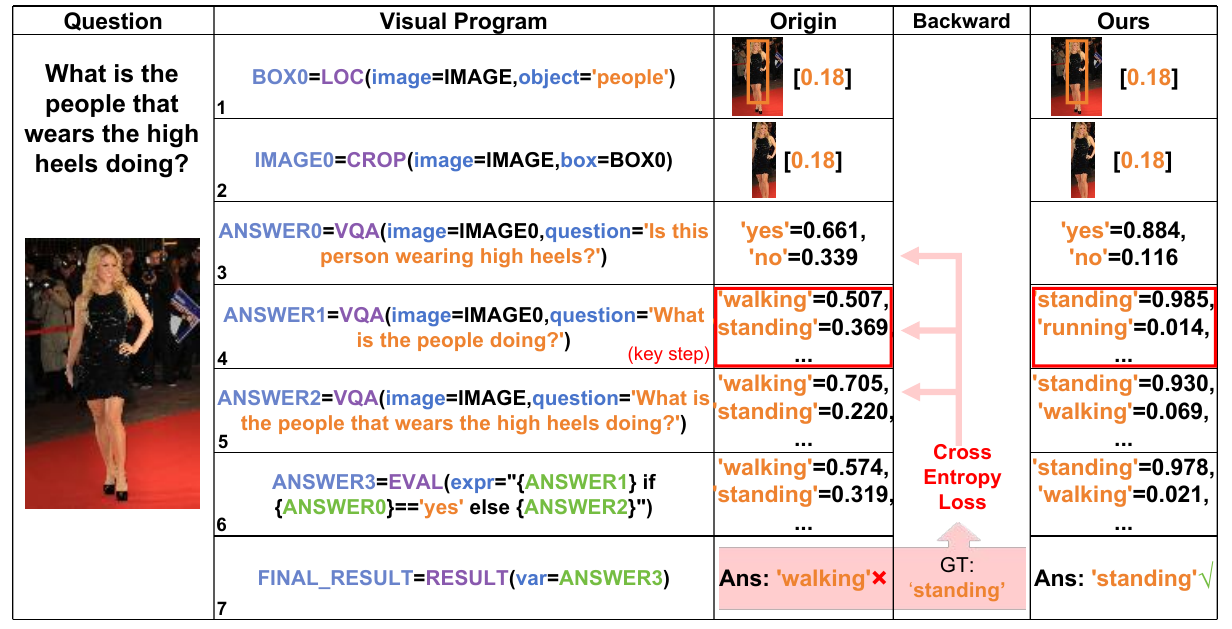}
    \caption{These program execution processes of the Visual Program before and after learning with our EVPG for a case in Open Images. We provide detailed process outputs of each step during program execution. `Ans' means the final answer, and `GT' means the ground truth.}
    \label{fig_case_combined_oi}
\end{figure}

In Fig.~\ref{fig_case_combined_gqa}, Fig.~\ref{fig_case_combined_nlvr}, and Fig.~\ref{fig_case_combined_oi}, we show more cases for our EVPG in the GQA~\cite{hudson2019gqa}, NLVRv2~\cite{suhr2019corpus}, and Open Images~\cite{shao2024visual} datasets.
Fig.~\ref{fig_case_combined_gqa} upper subgraph shows a case in GQA. This case asks ``Are there any fruits or toilets?'', and there are no fruits or toilets in the image.
So the ground truth is ``no''.
However, the LOC model in the first step makes a mistake in locating a bounding box of ``fruit''.
This mistake makes the final answer wrong.
After optimizing the LOC model using our EVPG, the mistake of the LOC model in the first step has been corrected.
The output of the LOC model in this step now produces no bounding box for the object ``fruit'', the final answer can also be corrected from ``yes'' to ``no''.
Fig.~\ref{fig_case_combined_gqa} lower subgraph shows a case in GQA, which is a shorter, similar case to the case shown in Fig.~\ref{fig_case}.

Fig.~\ref{fig_case_combined_nlvr} upper subgraph shows a case in NLVRv2. The statement is ``Each of the images features fresh watermelon slices along with popsicles''.
The pre-trained VQA model makes mistakes when answering ``Does the image feature popsicles?'' for both the left and right images.
Hence, the final answer is wrong, because there are no popsicles in both images.
After optimizing the VQA model using our EVPG, the outputs of the 1st and 2nd steps are still right, and the outputs of the 3rd and 4th steps are corrected: the probability of ``no'' is larger than ``yes''.
Then the final answer has been corrected from the original wrong answer ``True'' to the right answer ``False''.
Fig.~\ref{fig_case_combined_nlvr} lower subgraph shows another case in NLVRv2.

Fig.~\ref{fig_case_combined_oi} upper subgraph shows a case that asks ``What is the people who holds the microphone doing?'', the ground truth is ``sitting''.
However, the VQA model in the fourth step makes a mistake in determining what people are doing.
This mistake makes the final answer wrong.
After optimizing the VQA model using our EVPG, the mistake in the fourth step of the VQA model has been corrected.
The output of the VQA model in this step now produces the correct answer ``sitting'', the final answer can also be corrected from ``talking'' to ``sitting''.
Fig.~\ref{fig_case_combined_oi} lower subgraph shows another case in Open Images.

\subsection{Prompts for Visual Programs Generation}
\label{prompt_appendix}

We show the prompts used to generate the visual programs for GQA~\cite{hudson2019gqa}, NLVRv2~\cite{suhr2019corpus}, and Open Images~\cite{shao2024visual}.
Due to space limitations, only partial demonstrations are shown here for each Prompt. For the remaining demonstrations, please refer to the code.

\begin{figure}[t]
\centering
\footnotesize
\begin{tcolorbox}
Give the visual program to solve the problem. Only output the program. Do not output ```.
\\
You can only use these modules in the program: \\
BOX=LOC(image=IMAGE,object=``object'') \\
ANSWER=COUNT(box=BOX) \\
IMAGE=CROP(image=IMAGE,box=BOX) \\
IMAGE=CROP\_RIGHTOF(image=IMAGE,box=BOX) \\
IMAGE=CROP\_LEFTOF(image=IMAGE,box=BOX) \\
IMAGE=CROP\_INFRONTOF(image=IMAGE,box=BOX) \\
IMAGE=CROP\_BEHIND(image=IMAGE,box=BOX) \\
IMAGE=CROP\_BELOW(image=IMAGE,box=BOX) \\
IMAGE=CROP\_ABOVE(image=IMAGE,box=BOX) \\
ANSWER=VQA(image=IMAGE,question=``question'') \\
ANSWER=EVAL(expr=``expr'') \\
FINAL\_RESULT=RESULT(var=ANSWER) \\
Do not use any other modules! \\
\\
Question: Do the post and the sign have a different colors? \\
Program: \\
BOX0=LOC(image=IMAGE,object=``post'') \\
IMAGE0=CROP(image=IMAGE,box=BOX0) \\
BOX1=LOC(image=IMAGE,object=``sign'') \\
IMAGE1=CROP(image=IMAGE,box=BOX1) \\
ANSWER0=VQA(image=IMAGE0,question=``What color is the post?'') \\
ANSWER1=VQA(image=IMAGE1,question=``What color is the sign?'') \\
ANSWER2=EVAL(expr=```yes' if \{ANSWER0\} != \{ANSWER1\} else `no''') \\
FINAL\_RESULT=RESULT(var=ANSWER2) \\
\\
Question: What color is the curtain that is to the right of the mirror? \\
Program: \\
BOX0=LOC(image=IMAGE,object=``mirror'') \\
IMAGE0=CROP\_RIGHTOF(image=IMAGE,box=BOX0) \\
ANSWER0=VQA(image=IMAGE0,question=``What color is the curtain?'') \\
FINAL\_RESULT=RESULT(var=ANSWER0) \\
\\
Question: Does the traffic cone have white color? \\
Program: \\
BOX0=LOC(image=IMAGE,object=``traffic cone'') \\
IMAGE0=CROP(image=IMAGE,box=BOX0) \\
ANSWER0=VQA(image=IMAGE0,question=``Does the traffic cone have white color?'') \\
FINAL\_RESULT=RESULT(var=ANSWER0) \\
\\
Question: \textbf{\{new\_question\}} \\
Program:
\end{tcolorbox}
\caption{Prompt for DeepSeek-V3 to generate Visual Programs of GQA.}
\label{fig:gqa-prompt}
\end{figure}

\begin{figure}[t]
\centering
\footnotesize
\begin{tcolorbox}
Give the visual program to determine whether the statement is True or False. Only output the program. Do not output ```. Unless otherwise specified, the statement describes the combined situation in both images.
\\
You can only use these modules in the program: \\
ANSWER=VQA(image=IMAGE,question=``question'') \\
ANSWER=EVAL(expr=``expr'') \\
FINAL\_RESULT=RESULT(var=ANSWER) \\
Do not use any other modules! \\
\\
Statement: there are at least seven wine bottles in the image on the left \\
Program: \\
ANSWER0=VQA(image=LEFT,question=``How many wine bottles are in the image?'') \\
ANSWER1=EVAL(expr='\{ANSWER0\} >= 7') \\
FINAL\_ANSWER=RESULT(var=ANSWER1) \\
\\
Statement: One dog is laying down. \\
Program: \\
ANSWER0=VQA(image=LEFT,question=``How many dogs are laying down?'') \\
ANSWER1=VQA(image=RIGHT,question=``How many dogs are laying down?'') \\
ANSWER2=EVAL(expr=``\{ANSWER0\} + \{ANSWER1\} == 1'') \\
FINAL\_ANSWER=RESULT(var=ANSWER2) \\
\\
Statement: An image shows exactly two seals in direct contact, posed face to face. \\
Program: \\
ANSWER0=VQA(image=LEFT,question=``How many seals are in the image?'') \\
ANSWER1=VQA(image=RIGHT,question=``How many seals are in the image?'') \\
ANSWER2=VQA(image=LEFT,question=``Are the seals in direct contact?'') \\
ANSWER3=VQA(image=RIGHT,question=``Are the seals in direct contact?'') \\
ANSWER4=VQA(image=LEFT,question=``Are the seals posed face to face?'') \\
ANSWER5=VQA(image=RIGHT,question=``Are the seals posed face to face?'') \\
ANSWER6=EVAL(expr=``\{ANSWER0\} == 2 and \{ANSWER2\} and \{ANSWER4\}'') \\
ANSWER7=EVAL(expr=``\{ANSWER1\} == 2 and \{ANSWER3\} and \{ANSWER5\}'') \\
ANSWER8=EVAL(expr=``\{ANSWER6\} xor \{ANSWER7\}'') \\
FINAL\_ANSWER=RESULT(var=ANSWER8) \\
\\
Statement: \textbf{\{new\_statement\}} \\
Program:
\end{tcolorbox}
\caption{Prompt for DeepSeek-V3 to generate Visual Programs of NLVRv2.}
\label{fig:nlvr-prompt}
\end{figure}

\begin{figure}[t]
\centering
\footnotesize
\begin{tcolorbox}
Give the visual program to solve the problem. Only output the program. Do not output ```.
\\
You can only use these modules in the program: \\
BOX=LOC(image=IMAGE,object=``object'') \\
ANSWER=COUNT(box=BOX) \\
IMAGE=CROP(image=IMAGE,box=BOX) \\
IMAGE=CROP\_RIGHTOF(image=IMAGE,box=BOX) \\
IMAGE=CROP\_LEFTOF(image=IMAGE,box=BOX) \\
IMAGE=CROP\_INFRONTOF(image=IMAGE,box=BOX) \\
IMAGE=CROP\_BEHIND(image=IMAGE,box=BOX) \\
IMAGE=CROP\_BELOW(image=IMAGE,box=BOX) \\
IMAGE=CROP\_ABOVE(image=IMAGE,box=BOX) \\
ANSWER=VQA(image=IMAGE,question=``question'') \\
ANSWER=EVAL(expr=``expr'') \\
FINAL\_RESULT=RESULT(var=ANSWER) \\
Do not use any other modules! \\
\\
Question: What is the people that holds the guitar doing? \\
Program: \\
BOX0=LOC(image=IMAGE,object=``people'') \\
IMAGE0=CROP(image=IMAGE,box=BOX0) \\
ANSWER0=VQA(image=IMAGE0,question=``Is this person holding a guitar?'') \\
ANSWER1=VQA(image=IMAGE0,question=``What is the people doing?'') \\
ANSWER2=VQA(image=IMAGE,question=``What is the people that holds the guitar doing?'') \\
ANSWER3=EVAL(expr=``\{ANSWER1\} if \{ANSWER0\} == `yes' else \{ANSWER2\}'') \\
FINAL\_RESULT=RESULT(var=ANSWER3) \\
\\
Question: Who is on the plastic chair? \\
Program: \\
ANSWER0=VQA(image=IMAGE,question=``Who is on the plastic chair?'') \\
FINAL\_RESULT=RESULT(var=ANSWER0) \\
\\
Question: Where the standing boy is under? \\
Program: \\
BOX0=LOC(image=IMAGE,object=``standing boy'') \\
IMAGE0=CROP\_ABOVE(image=IMAGE,box=BOX0) \\
ANSWER0=VQA(image=IMAGE0,question=``Where the standing boy is under?'') \\
FINAL\_RESULT=RESULT(var=ANSWER0) \\
\\
Question: \textbf{\{new\_question\}} \\
Program:
\end{tcolorbox}
\caption{Prompt for DeepSeek-V3 to generate Visual Programs of Open Images.}
\label{fig:open-images-prompt}
\end{figure}
}

{
    \clearpage
    \small
    \bibliographystyle{IEEEtran}
    \bibliography{Reference}
}

\newpage

\section{Biography Section}


\begin{IEEEbiography}[{\includegraphics[width=1in,height=1.25in,clip,keepaspectratio]{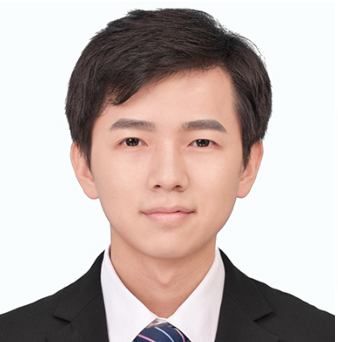}}]
{Wentao Wan} is currently a Ph.D candidate at the School of Computer Science and Engineering, Sun Yat-sen University, since 2019.
He received the M.S. degree in Computer Science and Technology from Huazhong University of Science and Technology in 2018, and the B.S. degree from Central South University in 2015.
His main research interests include Embodied AI, Agent, Multi-modal Understanding, General Reasoning, and Continual Learning.
\end{IEEEbiography}


\begin{IEEEbiography}[{\includegraphics[width=1in,height=1.25in,clip,keepaspectratio]{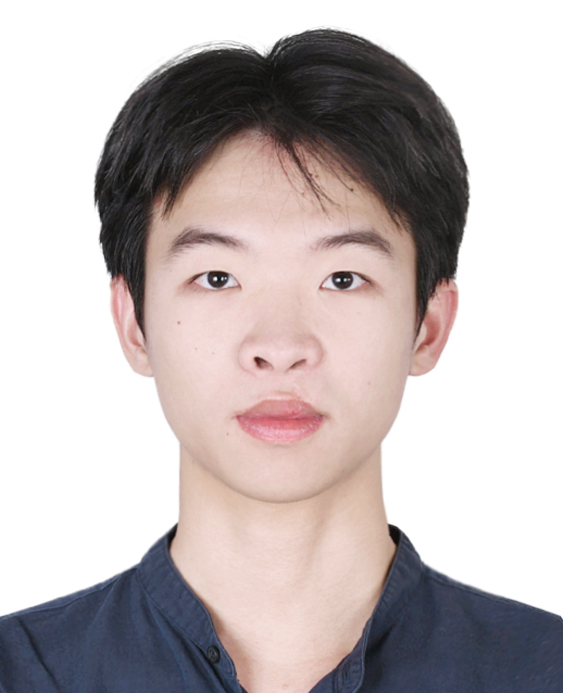}}]
{Kaiyu Wu} received the Bachelor’s degree in Software Engineering from Sun Yat-sen University, Zhuhai, China, in 2025.
He is currently a graduate student in progress at the School of Computer Science and Engineering, Sun Yat-sen University, Guangzhou, China.
His research interests include Reasoning Model, Agent, Reinforcement Learning, and Multi-modal Reasoning.
\end{IEEEbiography}


\begin{IEEEbiography}[{\includegraphics[width=1in,height=1.25in,clip,keepaspectratio]{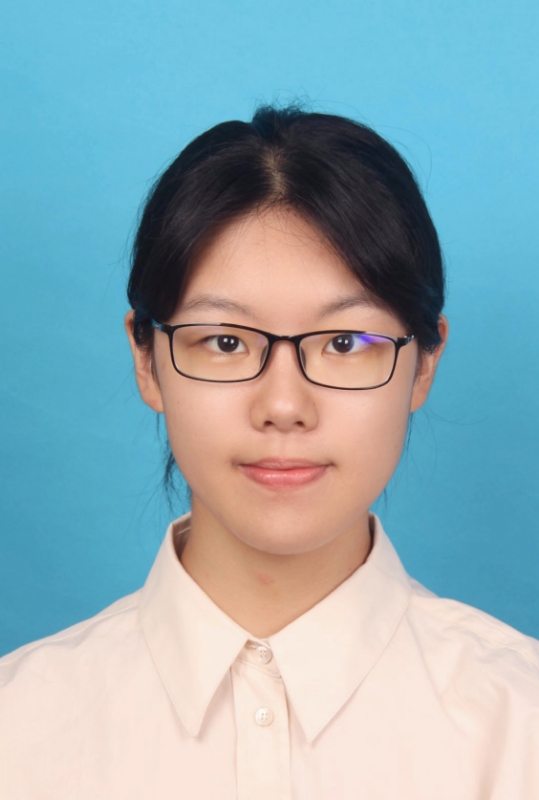}}]
{Qingyang Ma} is a final-year undergraduate student in Computer Science and Engineering at Sun Yat-sen University and a member of the HCP Lab.
Her research interests lie in Multi-modal Learning and the development of advanced AI methods for a diverse range of applications.
\end{IEEEbiography}


\begin{IEEEbiography}[{\includegraphics[width=1in,height=1.25in,clip,keepaspectratio]{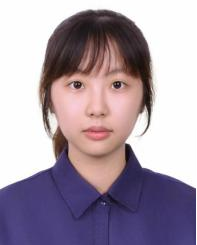}}]
{Nan Kang} received the Bachelor’s degree in Software Engineering from Central South University, Changsha, China, in 2024.
She is currently a graduate student in progress at the School of Computer Science and Engineering, Sun Yat-sen University, Guangzhou, China.
Her research interests include multi-modal reasoning, visual programming, and chain-of-thought.
\end{IEEEbiography}


\begin{IEEEbiography}[{\includegraphics[width=1in,height=1.25in,clip,keepaspectratio]{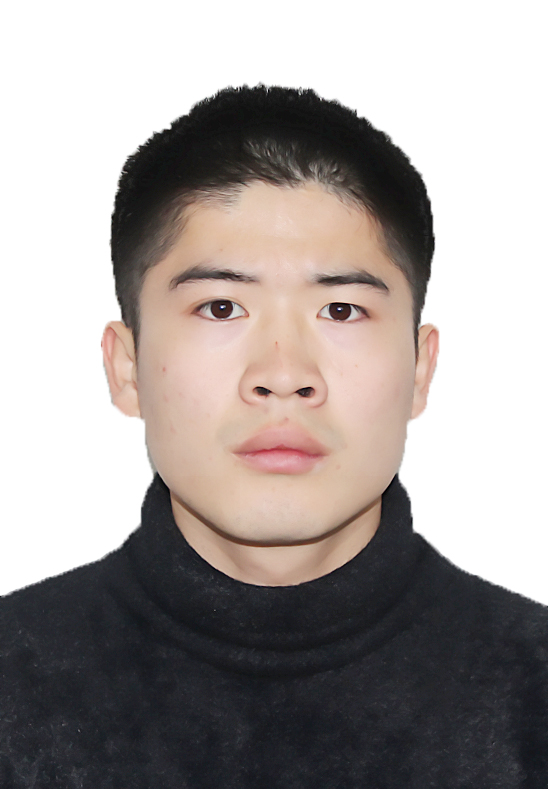}}]
{Yunjie Chen} received the B.Eng. degree from Wuhan University of Technology in 2024.
He is currently pursuing the M.Eng. degree in Computer Technology at Sun Yat-sen University.
His research interests include multimodal large language models and natural language processing.
\end{IEEEbiography}



\begin{IEEEbiography}[{\includegraphics[width=1in,height=1.25in,clip,keepaspectratio]{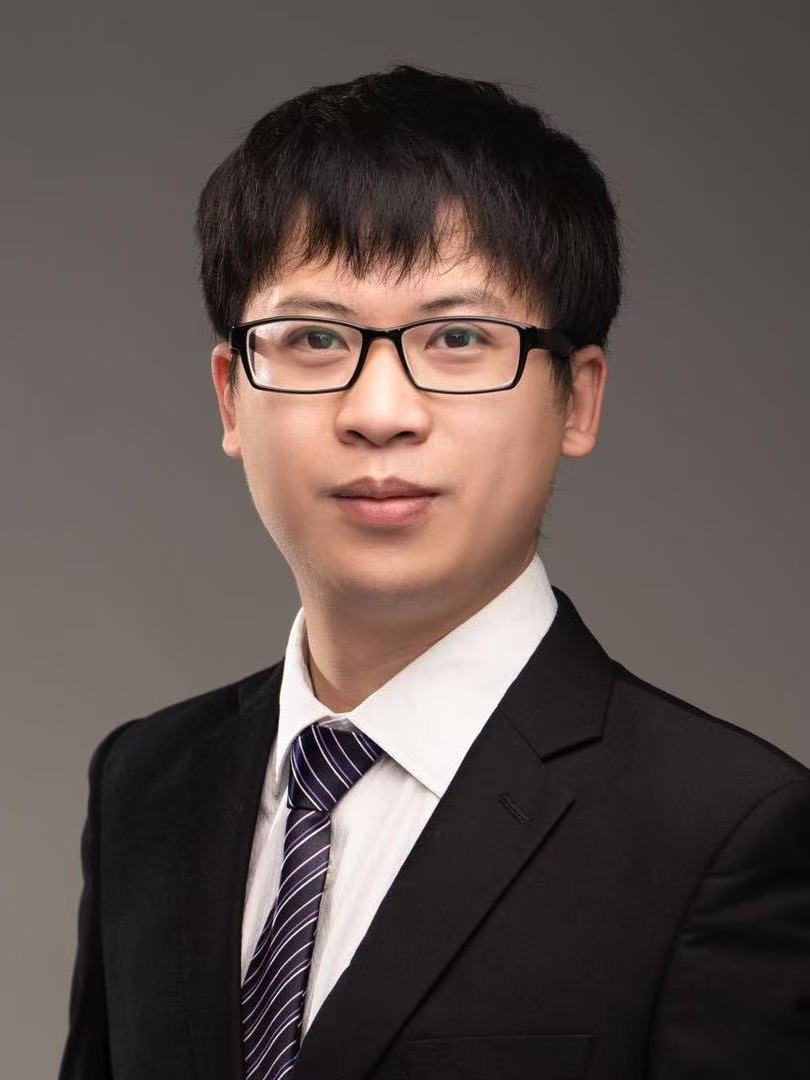}}]
{Keze Wang} is nationally recognized as the Distinguished Young Scholars, currently serving as an Associate Professor at the School of Computer Science, Sun Yat-sen University, and a doctoral supervisor.
He holds two Ph.D. degrees from Sun Yat-sen University (2017) and the Hong Kong Polytechnic University (2019).
In 2018, he worked as a postdoctoral researcher at the University of California, Los Angeles, and returned to Sun Yat-sen University in 2021 as part of the "Hundred Talents Program."
Dr. Wang has focused on reducing deep learning's dependence on training samples and mining valuable information from massive unlabeled data.
\end{IEEEbiography}

\begin{IEEEbiography}[{\includegraphics[width=1in,height=1.25in,clip,keepaspectratio]{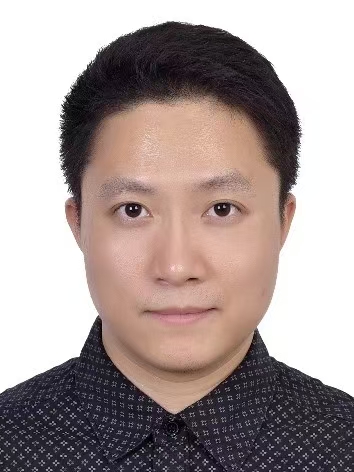}}]
{Liang Lin} (IEEE Fellow) is a full professor of computer science with Sun Yat-sen University.
He served as the executive director and distinguished scientist of SenseTime Group from 2016 to 2018, leading the R\&D teams for cutting-edge technology transferring. He has authored or co-authored more than 200 papers in leading academic journals and conferences, and his papers have been cited by more than 26\,000 times.
He is an associate editor of \textit{IEEE Trans. Neural Networks and Learning Systems} and \textit{IEEE Trans. Multimedia}, and served as area chairs for numerous conferences, such as CVPR, ICCV, SIGKDD, and AAAI.
He is the recipient of numerous awards and honors including Wu Wen-Jun Artificial Intelligence Award, the First Prize of China Society of Image and Graphics, ICCV Best Paper Nomination, in 2019, Annual Best Paper Award by Pattern Recognition (Elsevier), in 2018, Best Paper Dimond Award in IEEE ICME 2017, Google Faculty Award, in 2012.
His supervised PhD students received ACM China Doctoral Dissertation Award, CCF Best Doctoral Dissertation and CAAI Best Doctoral Dissertation. He is a fellow of IET/IAPR.

\end{IEEEbiography}

\vfill

\end{document}